%% file: main_arxiv.tex
\pdfoutput=1
\documentclass[lettersize,journal]{IEEEtran}
\IEEEoverridecommandlockouts
\usepackage{amsmath,amsfonts}
\usepackage{algorithmic}
\usepackage{algorithm}
\usepackage{array}
\usepackage[caption=false,font=normalsize,labelfont=sf,textfont=sf]{subfig}
\usepackage{textcomp}
\usepackage{stfloats}
\usepackage{url}
\usepackage{verbatim}
\usepackage{graphicx}
\usepackage{cite}
\usepackage{hyperref}
\usepackage{booktabs}
\usepackage{comment}
\usepackage{multirow}
\usepackage{subcaption}
\usepackage{subfig}
\usepackage{xcolor}
\usepackage[table]{xcolor}
\usepackage{amssymb}
\usepackage{listings}
\usepackage{hyperref}

\begin{document}
\bstctlcite{IEEEexample:BSTcontrol}
\title{LLM-guided Planar Loco-Manipulation for Quadruped Robots}

\author{Georges~Jetti$^{*}$, Michael~Khayyat, Jean-Pierre~Sleiman, Stefano~Arrigoni, and~Francesco~Braghin%
\thanks{G.~Jetti, M.~Khayyat, S.~Arrigoni, and F.~Braghin are with the Department of Mechanical Engineering, Politecnico di Milano, 20156 Milan, Italy.}%
\thanks{J.-P.~Sleiman was with the Robotics and AI Institute, USA, during this work. He is now with Flexion Robotics.}%
\thanks{$^{*}$Corresponding author: G.~Jetti (e-mail: georges.jetti@polimi.it).}%
}

\maketitle

\begin{abstract}

Robotic navigation typically assumes a collision-free route to the goal exists; in cluttered environments with movable obstacles, however, reaching the goal may require physical interaction. Classical solutions encode these behaviors through geometric rules and task-specific logic, while learning-based approaches often constrain the decision space to low-dimensional motion commands or predefined action sets. We instead consider a broader decision space, jointly selecting the skill, relevant object, and geometric target. Our language-conditioned task planner, fine-tuned with LoRA on synthetic data, maps natural-language tasks and a structured scene representation to variable-length sequences of parameterized actions, executed by low-level reinforcement-learning policies. The planner learns the geometric relations underlying interactive navigation and outperforms a classical planner and a prompted frontier model on interactive navigation tasks. Comparisons with models trained from scratch indicate that composition does not require pretraining when the representation is structured, whereas pretraining yields more reliable geometric targets. At the low level, we introduce an end-to-end whole-body planar pushing policy that achieves a 93.3\% success rate in simulation, versus 83.5\% for a hierarchical baseline, while reducing completion time by more than 60\%. On hardware, the hierarchy yields the correct skill composition in 44/45 trials and end-to-end success in 34/45. Project website: \url{https://anonymousprojects-dev.github.io/LLM-guided-LocoManipulation-Quadrupeds/}.

\end{abstract}

\section{Introduction}

Conventional robotic navigation pipelines typically %
assume the existence of a feasible path to the goal and that the robot's repertoire of navigation skills is sufficient to cope with structural constraints, such as large obstacles. In practice, these assumptions often fail. %
Instead, when no feasible collision-free path exists in the current environment configuration, the robot must reason about the scene and generate coordinated sequences of locomotion and manipulation actions to modify the environment and create a feasible route to the objective, analogous to how biological agents actively alter their surroundings to achieve task objectives. This class of problems is known as navigation among movable obstacles (NAMO) in the literature and, more broadly, interactive navigation, where navigation and manipulation actions must be coordinated to reach otherwise inaccessible goals~\cite{dai2024interactivenavigationadaptivenonprehensile,bi2025interactivenavigationleggedmanipulators,10341355,he2024interactivefarinteractivefastadaptablerouting,li2019hrl4inhierarchicalreinforcementlearning,yao2024localpathplanningpushable,zeng2021pushingwayinteractivevisual,yang2025efficientnavigationmovableobstacles,zhou2025adaptiveinteractivenavigationquadruped,11657384}.

Conventional NAMO approaches effectively combine geometric planning and explicit interaction logic~\cite{dai2024interactivenavigationadaptivenonprehensile,bi2025interactivenavigationleggedmanipulators,10341355,he2024interactivefarinteractivefastadaptablerouting}; however, accommodating different interaction cases generally requires extending the associated geometric rules, heuristics, or task-specific decision logic, increasing the engineering burden as the range of task configurations expands. Learning-based methods provide an alternative paradigm by acquiring interaction strategies directly from experience~\cite{li2019hrl4inhierarchicalreinforcementlearning,yao2024localpathplanningpushable,zeng2021pushingwayinteractivevisual,yang2025efficientnavigationmovableobstacles}. To manage the resulting learning problem, existing approaches adopt different task-specific abstractions of the decision space. The authors in~\cite{yao2024localpathplanningpushable} directly learn a policy over forward and angular velocities for both local navigation and interaction. Although effective, this formulation does not explicitly represent alternative manipulation modes, and the object interaction is induced implicitly through the velocity policy, providing less control over how the environment should be modified. Moreover, as investigated in Sec.~\ref{sec:HRLvsE2E}, coupling locomotion and planar manipulation through a common controller constrains the manipulation behavior, leading to lower success and higher task completion time under object variability. In~\cite{zeng2021pushingwayinteractivevisual}, the policy instead emits navigation and manipulation actions and learns to predict their effects on the surrounding objects; however, the action set has predefined geometric targets, limiting the range of interactions that can be performed. In~\cite{yang2025efficientnavigationmovableobstacles}, a hierarchical policy is employed in which the high-level controller generates a compact pushing command specifying the contact location, relative contact orientation, and base velocity. Nonetheless, the problem is formulated around following a preplanned global path, limiting the set of interactions with the environment. The method in~\cite{li2019hrl4inhierarchicalreinforcementlearning} jointly learns which segment of the robot to command and subsequently a continuous sub-goal. However, the evaluated tasks contain a limited set of task-level decisions: the relevant object to manipulate is effectively determined by the task structure rather than selected among multiple candidate objects based on their relation to the task objective and the environment. In contrast, we consider a planning problem in which the planner must jointly determine which skill to execute, which object to interact with (when applicable), and the geometric target associated with that skill. The cases considered in this work can admit multiple feasible task-level plans, while alternative choices of skill, object, or geometric target can result in infeasible plans. The planner must therefore reason over the relationship between the task objective and the scene geometry, rather than follow a prescribed interaction sequence. The resulting variable-length sequence of parameterized navigation and manipulation actions provides a flexible interface for composing different loco-manipulation strategies.%

Pretrained foundation models provide a natural means of reasoning over such variable task-level decisions. Prior work employing LLMs and VLMs to coordinate locomotion and manipulation skills for quadrupedal robots commonly relies on prompt engineering and in-context learning (ICL~\cite{brown2020languagemodelsfewshotlearners})~\cite{zhou2025adaptiveinteractivenavigationquadruped, ouyang2025longhorizonlocomotionmanipulationquadrupedal, qiu2025wildlmalonghorizonlocomanipulation,mei2024quadrupedgptversatilequadrupedagent,  chen2024commonsensereasoningleggedrobot,qu2026pragmatistrobotlearningplan,wang2025odyssey}. These approaches demonstrate that pretrained models can decompose long-horizon objectives, select among available skills, and, in some cases, infer continuous execution parameters. Nevertheless, the required planning behavior is elicited through task descriptions, skill specifications, and examples of how tasks can be solved, increasing the input context and inference latency during deployment. Moreover, pretrained LLMs are not inherently grounded in the robot-specific action space or operational constraints required for execution. Task-specific finetuning provides a means of adapting these pretrained representations to robotic planning and action generation~\cite{11657384,ravichandran2025distillingondevicelanguagemodels, kim2024openvlaopensourcevisionlanguageactionmodel, kim2025finetuningvisionlanguageactionmodelsoptimizing, nasrat2025rdmmfinetunedllmmodels, brohan2023rt2visionlanguageactionmodelstransfer, wei2026psi0openfoundationmodel, black2026pi0visionlanguageactionflowmodel, intelligence2025pi05visionlanguageactionmodelopenworld, nvidia2025gr00tn1openfoundation}, alleviating the need for large input prompts while grounding the model's outputs with respect to the robot's capabilities and environment. Task-specific finetuning, however, introduces its own costs: adapting large pretrained models can require updating a large number of parameters, while collecting sufficiently diverse expert demonstrations can be expensive.%

Closely related to our formulation,~\cite{11657384} fine-tune a Vision Language Model (VLM) for interactive navigation to determine when environmental interaction is required and to select the corresponding interaction skill and object. The geometric outcome of the interaction, however, is determined downstream by the skill-execution pipeline. Moreover, its counterfactual reasoning evaluates the removal of a candidate object, whereas in cluttered scenes the object must be relocated to a pose that itself remains feasible with respect to the surrounding objects. In contrast, our planner predicts the geometric target associated with each action. For manipulation, this explicitly specifies the desired object configuration, allowing the task planner to reason not only about which object to manipulate, but also where it should be moved. By decoupling the manipulation skill from a task-specific geometric outcome, this parameterization provides direct control over how the environment is modified and enables the same goal-conditioned manipulation skill to be reused across environment configurations.

The effectiveness of a high-level planner additionally depends on the capabilities of the low-level skills available to the robot. In particular, successful interactive navigation requires a manipulation policy that can reliably reposition obstacles while maintaining a stable behavior. Typically, quadruped manipulation is solved by mounting a multi-DOF arm on the quadruped body~\cite{sleiman2021unifiedmpcframeworkwholebody,doi:10.1126/scirobotics.adg5014}. An alternative paradigm for the manipulation of heavy objects is to leverage the base of the quadruped. Although limited to non-prehensile manipulation, it has become a viable option to push objects of various sizes, shapes, and masses both in a planar setting~\cite{jeon2023learningwholebodymanipulationquadrupedal,zhou2025adaptiveinteractivenavigationquadruped,ouyang2025longhorizonlocomotionmanipulationquadrupedal}, on slopes~\cite{sombolestan2023hierarchicaladaptivelocomanipulationcontrol}, and collaboratively with multiple robots~\cite{feng2025learningmultiagentlocomanipulationlonghorizon,sombolestan2023hierarchicaladaptivecontrolcollaborative}. Nevertheless, existing solutions constrain the manipulation controller by defining a specific gait cycle through a scheduler~\cite{sombolestan2023hierarchicaladaptivelocomanipulationcontrol,sombolestan2023hierarchicaladaptivecontrolcollaborative}, or more commonly by formulating a hierarchical loco-manipulation framework~\cite{jeon2023learningwholebodymanipulationquadrupedal,zhou2025adaptiveinteractivenavigationquadruped,ouyang2025longhorizonlocomotionmanipulationquadrupedal,feng2025learningmultiagentlocomanipulationlonghorizon} whereby a high-level planner feeds velocity commands to a fixed locomotion policy with prescribed gait, and body orientation. Such decoupling fails to produce an overall policy \textit{specialized for manipulation}; the locomotion-centric priors bias contact scheduling and base motion toward an optimal behavior for locomotion, not manipulation, thereby limiting efficiency and adaptability.%

\begin{figure*}[t]
\centering

    \includegraphics[width=\textwidth]{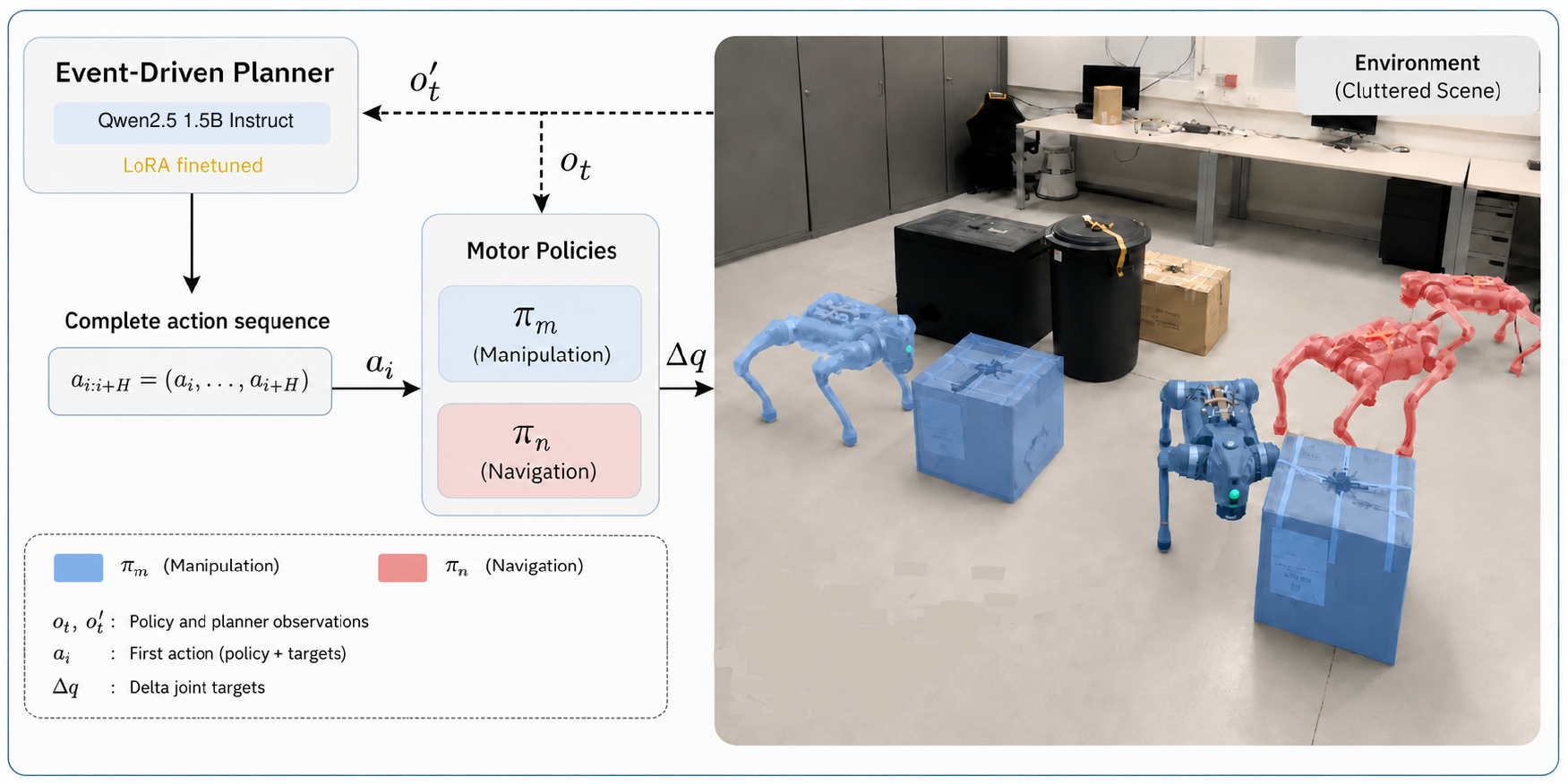}

\caption{Deployment pipeline of the proposed framework on the quadruped robot. The planner receives observation $o'_t$ consisting of a task instruction specified by the user, scene states such as robot and object pose, along with object parameters, and the previous skill that was applied on the environment. The planner, trained on predicting the remaining action sequence to achieve the task objective, applies only the first action in the sequence, and is queried again once that action is complete. Thus, our task planner is an \textit{event-driven task planner}. Blue robot instances signify that a manipulation policy was used in that stage, whereas Red signifies a navigation policy. Dashed lines represent data flow, whereas solid lines represent decision flow.}
\label{fig:framework}
\end{figure*}

To address the aforementioned limitations, we finetune a Qwen2.5-1.5B-Instruct model~\cite{qwen2025qwen25technicalreport} using LoRA~\cite{hu2021loralowrankadaptationlarge}. Inspired by recent work on robot learning~\cite{mandlekar2023mimicgendatagenerationscalable, jiang2025dexmimicgenautomateddatageneration,ravichandran2025distillingondevicelanguagemodels, nvidia2025gr00tn1openfoundation}, we finetune our planner on synthetic data. The proposed data generation pipeline enables rapid collection and curation of diverse scene configurations and demonstrations without requiring physical teleoperation~\cite{kim2024openvlaopensourcevisionlanguageactionmodel}, or continuous instantiation of a larger model to generate the trajectories~\cite{choudhury2024betterteacherllmagents,ravichandran2025distillingondevicelanguagemodels,nvidia2025gr00tn1openfoundation}. Given a natural-language task description and a structured geometric scene representation, the planner generates a variable-length sequence of parameterized navigation and manipulation actions, including the selected skill, object when applicable, and the associated geometric target for the selected skill. Additionally, unlike hierarchical whole-body planar manipulation frameworks~\cite{jeon2023learningwholebodymanipulationquadrupedal, ouyang2025longhorizonlocomotionmanipulationquadrupedal, zhou2025adaptiveinteractivenavigationquadruped}, we devise an \textit{end-to-end} manipulation policy that discovers an efficient planar pushing behavior. The resulting policy improves manipulation success while reducing task completion time by more than $60\%$ compared to the conventional hierarchical method. We evaluate the complete framework across simulated and real-world loco-manipulation tasks, including held-out in-distribution trajectories and a shifted task structure excluded from training that requires learned navigation and manipulation behaviors to be composed into a new multi-step strategy. Figure~\ref{fig:framework} illustrates the deployment of the proposed framework on the quadruped platform.

The main contributions of this work are summarized as follows:
\begin{itemize}
    \item \textbf{A language-conditioned task planner for interactive navigation}, finetuned with LoRA on synthetic demonstrations, that predicts the remaining sequence of parameterized actions by jointly selecting the skill, the relevant object, and the geometric target. The planner learns the geometric
    relations underlying interactive navigation without handcrafted rules or large prompts. Comparisons with models trained from scratch indicate that skill composition does not require pretraining when the representation is structured, whereas pretraining yields more reliable geometric targets.
    \item \textbf{An end-to-end whole-body planar manipulation policy} that enables specializing the robot's behavior to planar object pushing. The resulting policy improves manipulation success and reduces task-completion time by more than \(60\%\) relative to the common hierarchical manipulation baseline.
    \item \textbf{Repeated hardware evaluation of the complete framework}, coupling the language-conditioned planner with learned navigation and manipulation skills across diverse task structures, including the task structure excluded from training.
\end{itemize}

\vspace{-0.5em}
\section{Preliminaries}
\label{ss:Preliminaries}

For the planner, we assume access to robot and object pose observations, and 2D object geometry, thus focusing on the planning capabilities of the model. These observations can be obtained from a motion-capture system, an external vision system, or an onboard perception and mapping pipeline. In our hardware experiments, these quantities are provided by a motion-capture system. The manipulation policy has \textit{no prior knowledge} of the objects' dynamics or geometry.
We define a \textit{loco-manipulation} policy as one that is trained to both locomote to an object and manipulate it, while a \textit{manipulation} policy is one that can \textit{only} manipulate.

We consider two goal-conditioned motor skills for our low-level skill library: navigation ($s=n$) and planar manipulation ($s=m$). Each skill is modeled as a finite-horizon Markov Decision Process (MDP)
$
\mathcal{M}_s = (\mathcal{X}_s, \mathcal{A}_s, \mathcal{G}_s, p_s, r_s, \gamma);
$
where $\mathcal{X}_s$ and $\mathcal{A}_s$ denote the state and action spaces,
$p_s(x_{t+1} \mid x_t, a_t)$ is the transition dynamics,
$r_s(x_t, a_t, g_s)$ is the goal-conditioned reward with $g_s \in \mathcal{G}_s$,
and $\gamma \in (0,1)$ is the discount factor. A goal-conditioned policy parameterized by $\theta_s$, denoted by $\pi_{s}(a_t \mid x_t,  g_s)$ for simplicity, induces trajectories
$\tau = (x_0,a_0,\dots,x_T)$ under the dynamics $p_s$.
The optimal policy maximizes the expected discounted return:
\begin{equation}
\label{eq:ocp}
\pi_{s}^* \in \arg\max_{\pi_{s}}
\mathbb{E}_{\tau \sim p_s,\,\pi_{s}}
\left[
\sum_{t=0}^{T} \gamma^t r_s(x_t,a_t,g_s)
\right].
\end{equation}

Both low-level policies, $\pi_n$ and $\pi_m$, are  trained with Proximal Policy Optimization (PPO)~\cite{schulman2017proximalpolicyoptimizationalgorithms} in Isaac Lab~\cite{Mittal_2023} using domain randomization to enable zero-shot sim-to-real transfer.

At discrete high-level decision epochs $k$, the language-conditioned planner maps a natural-language task $G_{\mathrm{NL}}$ and structured scene representation to a variable-length sequence of parameterized actions denoted by $\mathbf{a^{HL}}_{k:k+H_k}$ where,
\begin{equation}
    \mathbf{a^{HL}}_{k:k+H_k}
    =
    \left(
    a^{HL}_k,
    a^{HL}_{k+1},
    \ldots,
    a^{HL}_{k+H_k}
    \right),
    \qquad
    a^{HL}_j = (s_j, i_j, g_j)
\end{equation}
where $H_k$ denotes the predicted remaining skill horizon and $j \in \{k,\ldots,k+H_k\}$ indexes the predicted skills. Here, $s_j \in \{n,m\}$ denotes the selected skill, $i_j$ the relevant object identifier when applicable, and $g_j$ its geometric target. Although the planner predicts the complete remaining action sequence $\mathbf{a^{HL}}_{k:k+H_k}$, only $a^{HL}_k$ is executed before the planner is queried again at the next high-level decision using the updated scene state.%

\section{Methodology}
\label{ss:methodology}

\subsection{Planar Manipulation Policy}
\label{ss:mani_policy}

\paragraph{Observation and Action space}  We use an asymmetric actor-critic architecture~\cite{pinto2017asymmetricactorcriticimagebased}. The actor receives observations $\mathbf{o_t=(x_t,g_t)}$, whereas the critic additionally receives noiseless, privileged information $\mathbf{o_t^{\text{pr}}}$ that is unavailable to the robot in deployment allowing it to better estimate the value function under object diversity. The list of observations is documented in Table~\ref{tab:observations}. The action is a vector of desired joint-angle offsets $\Delta\mathbf{q}_{\text{des}}\in\mathbb{R}^{12}$, tracked by a joint-space impedance controller to generate torques $\boldsymbol{\tau}$.

\begin{table}[t]
\centering
\caption{Observations for the actor ($o_m$) and critic ($o_m, o_m^{\text{pr}}$) for the manipulation policy.}
\label{tab:observations}
\renewcommand{\arraystretch}{1.2}
\resizebox{\linewidth}{!}{\begin{tabular}{clcc}
\toprule
\textbf{Quantity} & \textbf{Description} & \textbf{Dim} & \textbf{Noise} \\
\midrule
\multicolumn{4}{c}{$\mathbf{o_t}$} \\
\midrule
${}^{R}\!\mathbf{p}_{RG}$ & Goal-Robot relative position w.r.t. Robot & 3 & -- \\
$\!\mathbf{\psi}_{RG}$&Goal-Robot relative heading & 1 & -- \\
$\Delta \mathbf{q}_j$&Joint position offsets from default & 12 & $\mathcal{U}(-0.01, 0.01)$ \\
$\Delta \mathbf{\Dot{q}}_j$&Joint velocities from default & 12 & $\mathcal{U}(-1.0, 1.0)$ \\
$\mathbf{a}_{t-1}$&Previous actions & 12 & -- \\
$\boldsymbol{\omega}_{R}$&Robot base angular velocity & 3 & $\mathcal{U}(-0.10, 0.10)$ \\
$\! \lVert\mathbf{p}_{OG} \rVert$&Goal-Object distance & 1 & $\mathcal{U}(-0.05, 0.05)$ \\
$\!\mathbf{\psi}_{GO}$&Object-Goal relative heading & 1 & $\mathcal{U}(-0.02, 0.02)$ \\
${}^{R}\!\mathbf{p}_{RO}$&Object-Robot relative position w.r.t Robot & 3 & $\mathcal{U}(-0.05, 0.05)$ \\
$\!\mathbf{\psi}_{RO}$&Object-Robot relative heading & 1 & $\mathcal{U}(-0.02, 0.02)$ \\
${}^{R}\!\mathbf{q}_{RO}$ & Object-Robot relative orientation (quaternion) w.r.t Robot & 4 & $\mathcal{U}(-0.05, 0.05)$ \\
\midrule
\multicolumn{4}{c}{$o_t^{\text{pr}}$} \\
\midrule
$\lambda$&Robot--object contact state & 1 & -- \\
$\mathbf{v}_{R}$ & Robot base linear velocity & 3 & --\\
$\mathbf{g_{R}}$ & Projected Gravity on the Robot frame & 3 & -- \\
${}^{R}\!\mathbf{p}_{COM}$& Object CoM position w.r.t. Robot  & 3 & -- \\
$m$& Object mass & 1 & -- \\
$\mathbf{d}$& Object bounding-box dimensions & 3 & -- \\
$\mathbf{I}_{O}$& Object principal moments of inertia & 3 & -- \\
$\!\mathbf{v}_{O}$ & Object linear velocity & 3 & -- \\
$\!\boldsymbol{\omega}_{O}$& Object angular velocity& 3 & -- \\
$\boldsymbol{\kappa}_{T}$& Object type & 1 & -- \\
$\boldsymbol{\mu}_{sd}$&Object dynamic and static friction coefficients & 2 & -- \\

\bottomrule
\end{tabular}}
\end{table}

\paragraph{Rewards}

To avoid weight-tuning between translation and yaw errors, we define a keypoint set $\mathcal{K}$ on the object, similar to~\cite{jeon2023learningwholebodymanipulationquadrupedal, petrenko2023dexpbtscalingdexterousmanipulation, Allshire_2022}, to embed both errors into the same space. %
Let ${}^{W}\!\mathbf{p}_O^{(i)}$, ${}^{W}\!\mathbf{p}_G^{(i)}$ be the world positions of keypoint $i$ on the current object pose and the goal pose, respectively. The primary task reward is

\begin{equation}
rew_{\text{kp}} = \exp\left(-\sum_{i \in \mathcal{K}} \ell_{\text{kp}}^i / \sigma_{\text{kp}}^2 \right)
\end{equation}
where
\begin{equation}
\ell_{\text{kp}}^i = \left\| {}^{W}\!\mathbf{p}_O^{(i)} - {}^{W}\!\mathbf{p}_G^{(i)} \right\|^2
\end{equation}

To guide early learning over long distances, we formulate a velocity command $\mathbf{v}_{cmd}$, correlated to the target vector ${}^{R}\!\mathbf{p}_{OG}$, and reward for small error between object velocity ${\mathbf{v}}_{O}$ and $\mathbf{v}_{cmd}$. Previous planar manipulation work~\cite{jeon2023learningwholebodymanipulationquadrupedal,Dadiotis_2025} reward for object motion inline with the target vector. Although our reward adds a tuning parameter $K$%
, we empirically find that it offers better guidance for the agent, and emerges with a  behavior that gradually decreases object velocity throughout the task. The velocity command is formulated as follows:

\begin{equation}
rew_v = \exp\!\left(-\frac{\lVert \mathbf{v}_{\text{cmd}} - \mathbf{v}_O \rVert^{2}}{\sigma^{2}}\right)
\end{equation}
where
\begin{equation}
\mathbf{v}_{\text{cmd}} = {}^{R}\hat{\!\mathbf{p}}_{OG}\; \operatorname{min}\!\left(
K \lVert {}^{R}\!\mathbf{p}_{OG}\rVert, \ 0.75 \right)
\end{equation}

We find that including a reward for positive robot linear velocity along its x-axis , i.e. ${v}_{R,x}>0$, improves the stability of the robot.
To obtain a smooth behavior, we penalize joint acceleration $\mathbf{\Ddot{q}}$, joint torques $\boldsymbol{\tau}$, action rate $\Delta \mathbf{a_{t-1,t}}$, and object roll/pitch velocities $\boldsymbol{\omega}_{O,xy}$. We also find that penalizing feet acceleration helps reduce impact forces with the ground, decreasing the energy expenditure throughout the task. The list of rewards and their weights is provided in Appendix~\ref{sec:manipulationA}.%

\paragraph{Training Details}

We train the policy on %
cubes and cylinders, with randomized physical and geometric parameters%
, as detailed in Table~\ref{tab:training_params}. At the beginning of each episode, we randomize robot yaw and goal pose to enhance generalization. Object pose is randomized; however, we ensure small initial distance between the robot and the object. %
To further promote robustness, external disturbances are applied to the robot base at randomized intervals during training, implemented as base twist perturbations. Episode termination occurs under any of the following conditions: \textit{(i)} object flip over (i.e. the roll/pitch exceed a predefined threshold), \textit{(ii)} robot flip over, or \textit{(iii)} Episode Timeout (not failure). %
\begin{table}[t]
\centering
\caption{Training parameters in simulation.}
\label{tab:training_params}
\renewcommand{\arraystretch}{1.2}
\begin{tabular}{lcc}
\toprule
\textbf{Parameters} & \textbf{Training Range} \\
\midrule
Mass $(\mathrm{kg})$ & $\mathcal{U}(1,\,6)$  \\
$\Delta$ CoM position (cm) & $\mathcal{U}(-10.0,\,10.0)$ \\
Static Friction & $\mathcal{U}(0.5,\,1.2)$ \\
Dynamic Friction & $\mathcal{U}(0.6,\,1.0)$ \\
Object type & [Box, Cylinder] \\
Object diameter $(\mathrm{m})$ &  $\mathcal{U}(0.3,\,0.6)$  \\
Object width $(\mathrm{m})$ &  $\mathcal{U}(0.3,\,0.6)$ \\
Object depth $(\mathrm{m})$ &  $\mathcal{U}(0.3,\,0.6)$\\
Object height $(\mathrm{m})$ &  $\mathcal{U}(0.3,\,0.6)$ \\
\bottomrule
\end{tabular}
\end{table}

\subsection{Navigation Policy}
\label{ss:loco_policy}

We train a hierarchical navigation policy $\pi_n$ to track planar pose commands $\mathbf{c_t} \in \mathbb{R}^3$ (x, y, yaw) expressed in the base frame of the robot, while avoiding obstacles. We adopt a navigation policy rather than a pure point-to-point locomotion policy, as this removes the need for the LLM to generate intermediate locomotion waypoints, thereby reducing output sequence length and simplifying action generation. The low-level policy is trained to track velocity commands ($v_x$, $v_y$, $\psi$) and outputs desired joint angles w.r.t the default configuration, $\Delta q_{\text{desired}}$. %
For additional details on the navigation policy, please refer to the Appendix~\ref{sec:navigationA}.
The main task rewards for the navigation planner are expressed as follows:
\begin{equation}
rew_t = \exp\!\left(-\frac{\lVert {}^{W}\textbf{p}_{G}- {}^{W}\textbf{p}_R\rVert^2}{\sigma_t^2}\right)
\end{equation}
\begin{equation}
~rew_r = \exp\!\left(-\frac{\lVert {\psi}_{\text{G}} - {\psi}_R\lVert ^2}{\sigma_r^2}\right)
\end{equation}

\subsection{High-level LLM Task Planner}
\label{ss:LLM_planner}

\paragraph{Data Generation and Augmentation Strategy}
To systematically structure the training distribution, we introduce a taxonomy of task cases that partitions the dataset into mutually exclusive categories, ensuring systematic coverage of the problem space. These cases correspond directly to the task families summarized in Table~\ref{tab:planner_task_families}. For robot-centric objectives, we define the following cases:

\begin{itemize}
    \item \textbf{Case 1:} No objects in the scene.
    \item \textbf{Case 2:} One or more objects are present but do not obstruct the robot's path to the goal.
    \item \textbf{Case 3:} Two or more objects are arranged such that they obstruct direct navigation to the goal.
\end{itemize}

Cases 1 and 2 correspond to the \emph{Direct Navigation} task family in Table~\ref{tab:planner_task_families}, and can be completed through direct navigation. We additionally define an object-centric case, denoted \textbf{Case 2.5}, in which the robot is randomly positioned away from a single target object and must first navigate toward it before manipulation. The goal of this task is to facilitate the learning of the underlying manipulation skill. This corresponds to the \emph{Object Manipulation} task family. Case 3 provides the scene structure underlying the interactive task families, where successful completion requires manipulation and navigation to be combined. Depending on the initial robot-object configuration, the training data contains \emph{Interactive Navigation} trajectories with composition $M \rightarrow N$ and \emph{Long-Horizon Interactive Navigation} trajectories with composition $N \rightarrow M \rightarrow N$.
All data is generated in a custom 2D simulation environment implemented in Pygame, where the scene is a room of dimensions $3\times5$ [$\text{m}^2$] emulating the space in the lab. Scenes are procedurally constructed and evaluated at discrete state transitions, where each action deterministically updates the environment state.
We record scene states at each discrete decision step (i.e., skill-level transitions). This yields an event-driven trajectory representation
\[
\mathcal{D}_i = \{(z_{i,0}, a^{HL}_{i,0}),\dots, (z_{i,k}, a^{HL}_{i,k}), \dots, (z_{i,N_k}, a^{HL}_{i,N_k})\},
\]
where $i$ denotes the trajectory index and $N_k$ is the number of skill transitions in trajectory $i$. Here $z_{i,k}$ represents the symbolic scene states and parameters at decision step $k$, and $a^{HL}_{i,k}$ denotes the selected high-level action. Each scene encodes the robot pose, object poses, object geometry, a binary flag indicating object mobility, and a unique identifier assigned to each object. The number of manually-generated trajectories is dependent on the difficulty of the introduced cases. For instance, cases that require skill switching require more emphasis on manually generated data to ensure adequate coverage of the state-action pairs. Once the manual generation phase is done, we augment the trajectories using a set of simple rules to automate the enrichment of the dataset. For example, we mirror the scene and goal w.r.t each axis, we randomize robot, goal and object pose (and identifier), we insert irrelevant objects, etc.

This multi-stage data generation combines manual structural design with controlled stochastic augmentation, enabling scalable dataset expansion while preserving the semantic integrity of each task case. After augmentation, the resulting dataset contains approximately 8K trajectories, corresponding to around 16K training samples. For additional information on data generation, and augmentation, we refer the reader to Appendix~\ref{sec:DatagenerationA}.

\paragraph{Model Finetuning}
\label{ss:finetuning}
We fine-tune the Qwen2.5-1.5B-Instruct model, a decoder-only causal transformer
architecture, using Low-Rank Adaptation (LoRA). %
We fine-tune the model using standard cross-entropy loss for next-token prediction in a causal language modeling setting. Given an input sequence consisting of (i) the task instruction %
, (ii) the symbolic scene states and parameters $z_{k,j}$, including robot pose and object attributes, (iii) the skill executed at the previous step, and (iv) a \texttt{Done} flag indicating trajectory termination, the model auto-regressively predicts a structured action output. %
Rather than predicting only the immediate next high-level action, we train the model to predict a sequence of high-level actions within the trajectory, commonly referred to as action chunking (AC)~\cite{zhao2023learningfinegrainedbimanualmanipulation}, applied in a receding-horizon manner. In our case, the target is the remaining action sequence to reach the goal. %
Formally, we model
\[
p\left(a_{i,k:k+H} \mid z_{i,k}\right)
\]
where $H$ denotes the remaining skill horizon in our case. LoRA is applied to all linear layers. The LoRA rank ($r = 128$) is selected on a validation set of shifted-task scenes that is disjoint from the test scenes reported in Section~\ref{sec:Sim_comparison_llm}. We use a learning rate of $\eta = 2 \times 10^{-5}$. Additional information on finetuning and our design choices are detailed in Appendix~\ref{sec:LLMfinetuningA}.  %
\section{Experiments and Results}
\label{ss:experiments and results}

We evaluate the proposed framework at both levels of the hierarchy. Low-level manipulation is evaluated in physics simulation and through repeated hardware trials to assess task performance and sim-to-real transfer. High-level planning is evaluated separately in the procedural geometric environment described in~\ref{ss:LLM_planner}. Finally, repeated hardware experiments evaluate the complete hierarchy by coupling the pretrained planner with the learned navigation and manipulation policies. We use the Unitree Go2 robot in our experiments.

In this section, all end-to-end (E2E) policies run at 50 Hz while the impedance controller at the joints runs at 200 Hz. The planner of the hierarchical (HRL) policies runs at 10 Hz, while its low-level policy and joint controller run at 50 Hz and 200 Hz, respectively. All real-world experiments reported in this work are inferred on an RTX 5080 GPU, where LLM inference runs at approximately 1 Hz. Additional information on training is provided in Appendix~\ref{sec:TrainingA}.

\subsection{Preliminary Experiments}
\label{sec:HRLvsE2E}
\begin{figure}[t]
    \centering
    \subfloat[\label{fig:HRL_gait}]{%
      \includegraphics[width=0.5\linewidth]{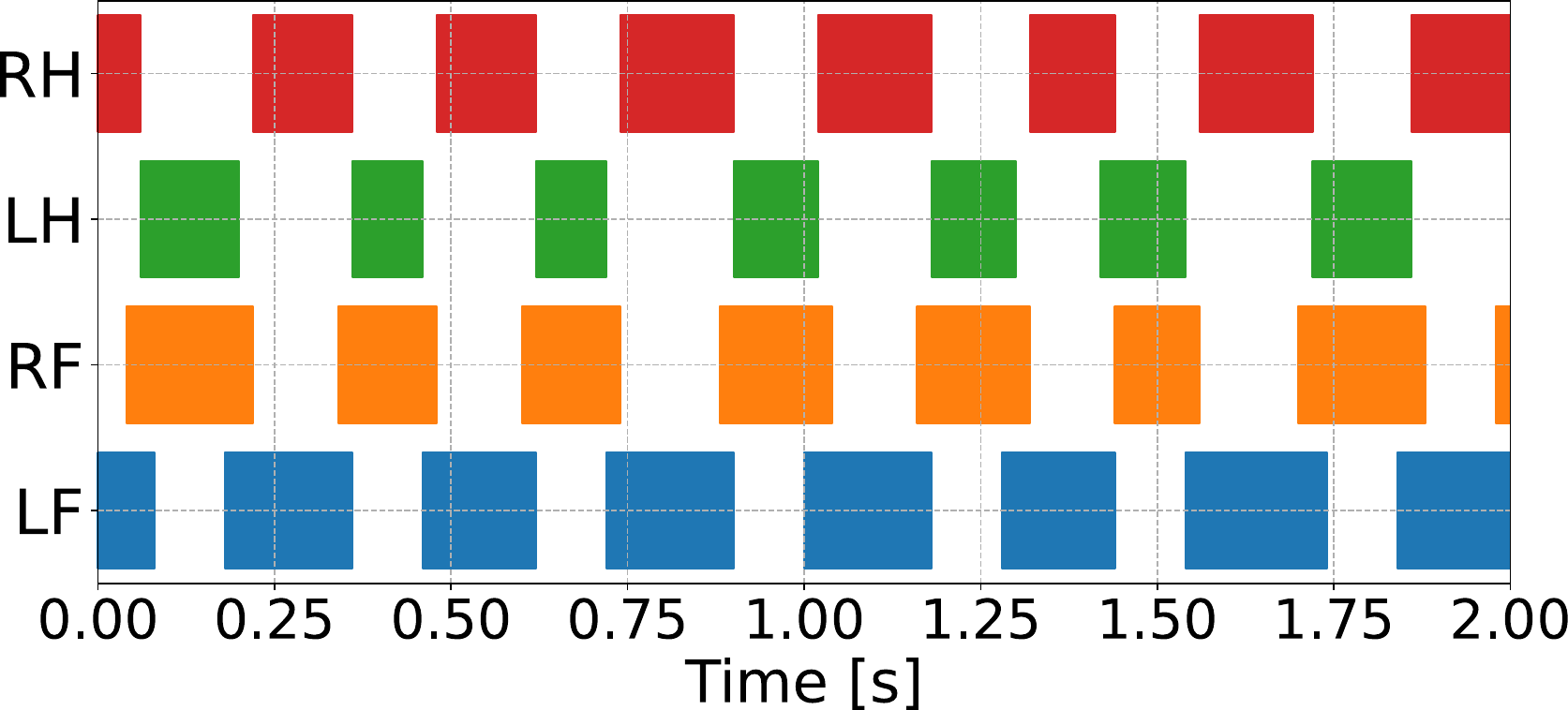}}
    \hfill
    \subfloat[\label{fig:E2E_gait}]{%
       \includegraphics[width=0.5\linewidth]{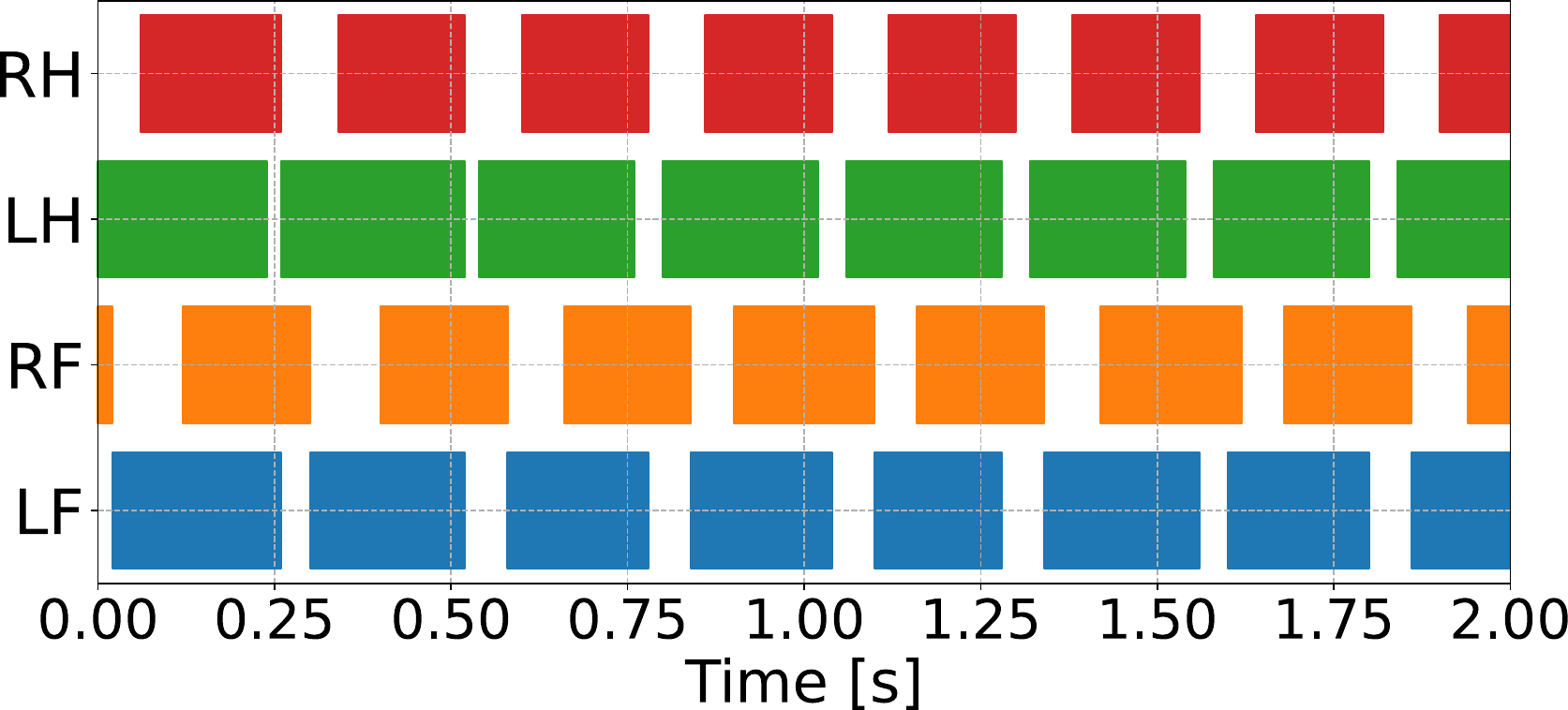}}
    \caption{\centering Emergent Gait Cycle for \textbf{(a) HRL}, and \textbf{(b) E2E} Manipulation Policies. Results were obtained from simulation.}
    \label{fig:emergent_gait}
\end{figure}

\begin{figure}[t]
\centering
\subfloat[\label{fig:roll}]{%
\href{https://youtu.be/KKf--W0IBuU}{%
\includegraphics[width=0.5\linewidth]{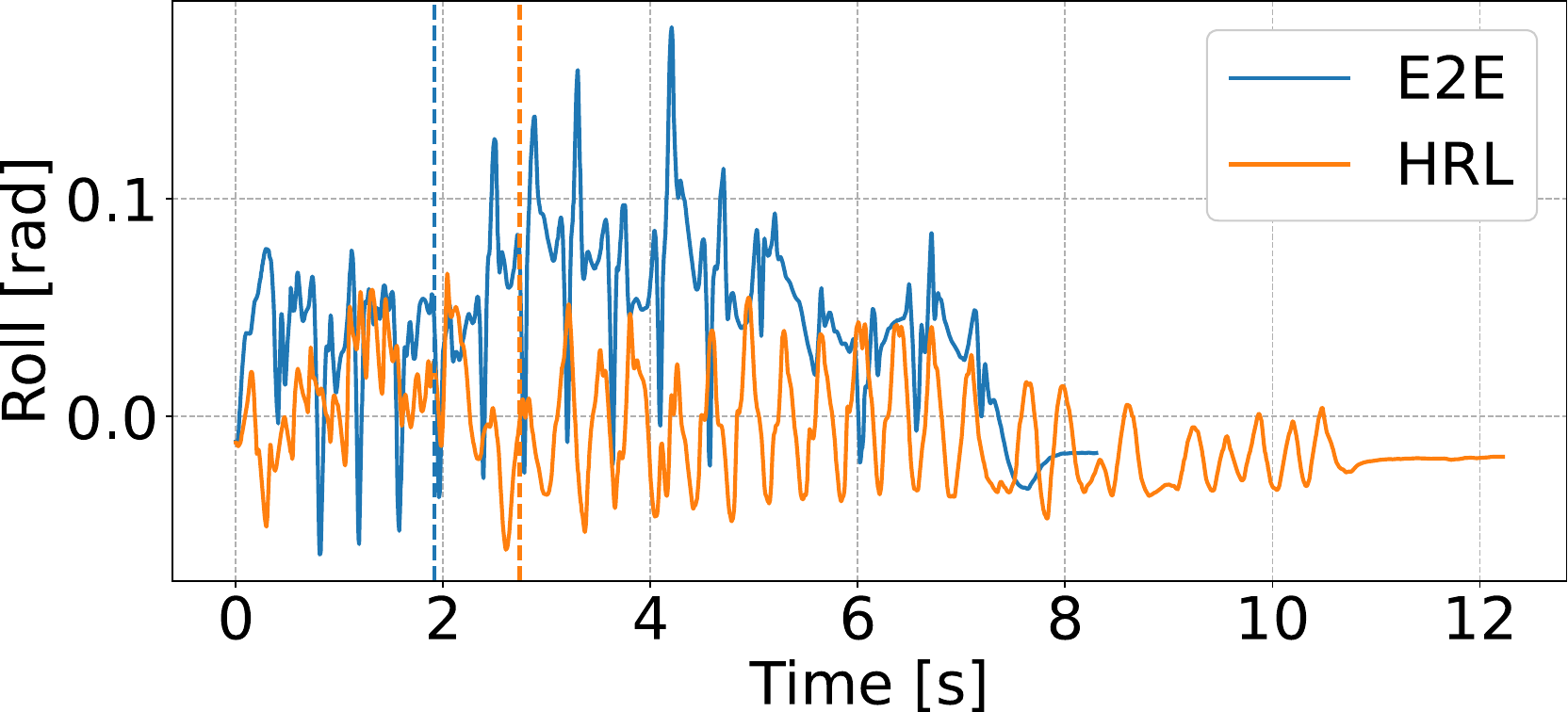}}}
\hfill
\subfloat[\label{fig:pitch}]{%
\includegraphics[width=0.5\linewidth]{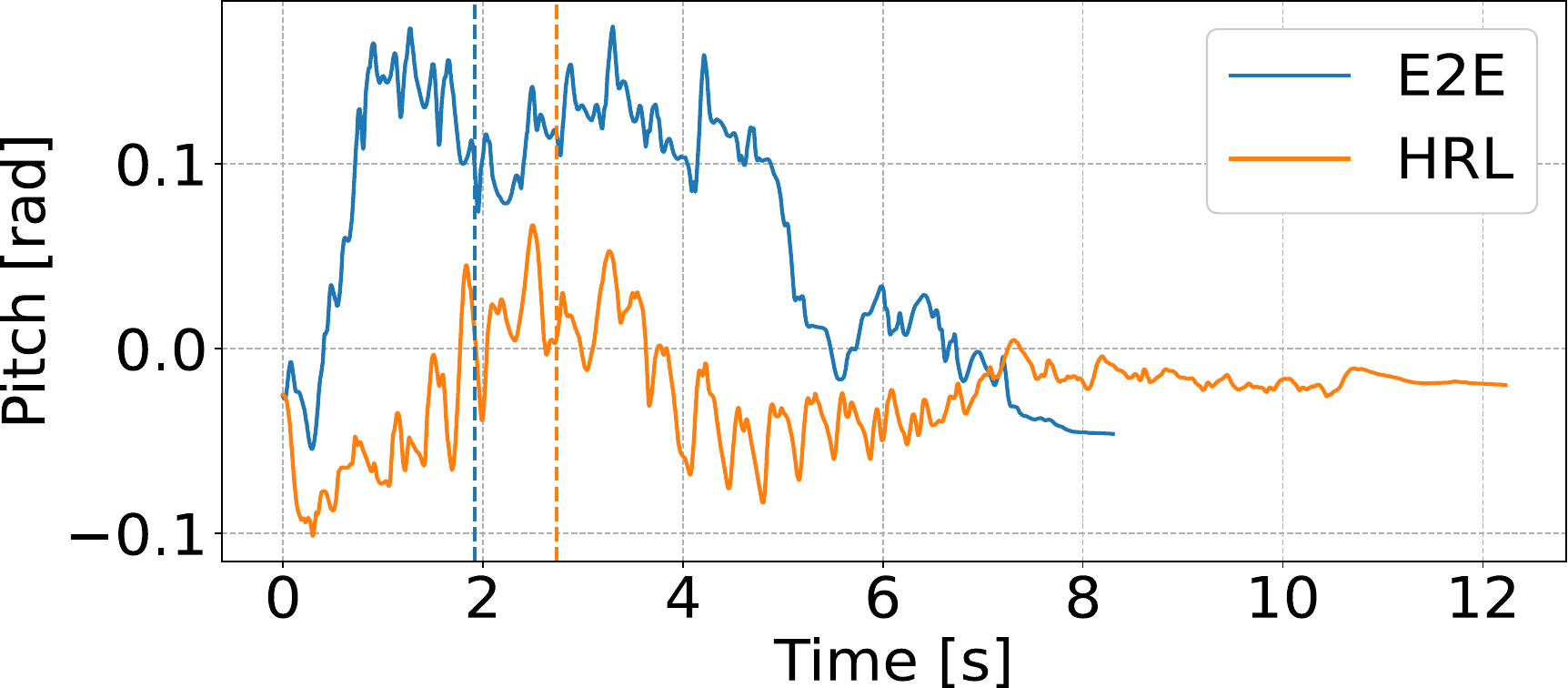}}
\caption{Comparison of body orientation, \textbf{(a) Roll}, and \textbf{(b) Pitch} for a run where the robot is distant from the object. Blue and orange dashed lines denote the first instant of contact with the object for E2E and HRL, respectively. The manipulate object is a box with an edge length of 40 [cm].}
\label{fig:body_orientation_comparison}
\end{figure}
We begin by demonstrating a set of experiments comparing HRL and E2E loco-manipulation policies, which motivates the design of our low-level skill library. To properly demonstrate the differences between these two policies, and ensure a fair comparison, we train both loco-manipulation policies from scratch under the same environment configuration.
To train the HRL loco-manipulation policy, we follow the pipeline of~\cite{jeon2023learningwholebodymanipulationquadrupedal}, excluding the adaptation module, to isolate the influence of the control architecture. The E2E policy, also without the adaptation module, is composed of a single MLP. To train that policy, we first reward the robot for reaching the object and activate the manipulation reward after $L$ training steps. For additional details on the training of these policies, we refer the reader to Appendix~\ref{sec:HRLvsE2EA}.
Figure~\ref{fig:emergent_gait} shows the emergent gait cycle for HRL and E2E loco-manipulation policies, respectively. Although the difference in training scheme may seem minimal, the HRL policy is intrinsically limited by the locomotion priors of the velocity-tracking policy. Thus, the planner of the HRL policy, the only module that has access to object pose, is optimized under a constrained setting. Instead, we see that the E2E policy learns an asynchronous gait cycle, and exhibits more base oscillations  as seen in figure~\ref{fig:body_orientation_comparison}, since it is free to optimize its entire behavior in order to maximize the expected rewards. The emergent gait cycle can be connected to the emergent body-orientation of the robot. The robot pitches its body, shifting its weight to the front. This shift in configuration makes it harder to lift the hind feet without losing stability. The emergent behavior allows the robot to position more of its inertia into the object allowing for more efficient transfer of momentum, and the lower point of contact aligns better with the CoM of the object. %
One caveat of the E2E policy is that it consumes more power. The mean power consumption for the E2E policy is 127.23 [W] compared to 66.03 [W] for the HRL policy.

While the emergent behavior of the E2E policy is effective for object-centric tasks, it is unnecessary for locomotion: increased base oscillations raise the probability of falling over, while the overall behavior results in higher energy consumption. We provide additional discussions and experiments on energy efficiency in Appendix~\ref{sec:HRLvsE2EA}.

\textit{In summary}, these results motivate our decomposition of planar loco-manipulation into specialized navigation and manipulation policies, where each segment can be independently solved by its own specialized policy instead of optimizing for both objectives in a single policy, which can lead to sub-optimal results~\cite{pmlr-v119-xu20h,munn2025scalablemultiobjectiverobotreinforcement}. Additionally, as we will see in Sec.~\ref{ss:Baselines_mani}, since the E2E policy is not constrained by any priors, and exhibits a broader range of adaptation towards diverse objects.

\subsection{Low-Level Manipulation Evaluation}
In this section, we present the results of our low-level manipulation policy in simulation and on hardware. Each baseline follows its corresponding training procedure described in~\ref{sec:HRLvsE2E}, with the exception of initializing the robot near the object for the HRL manipulation policy. All policies in this section, including ours, share the same observations, object distribution, and domain randomization (Table~\ref{tab:training_params}). We tune the reward weights of each policy to ensure a fair comparison.

Additionally, we ablate two reward terms from our proposed setting in order to better understand their effect on planar manipulation and provide the results in Appendix~\ref{sec:Ablationmani}.

\label{ss:Baselines_mani}
\subsubsection{Simulation Comparison}
The proposed policy is compared against the following baselines: \textit{(1)} E2E policy trained for loco-manipulation, and \textit{(2)} HRL policy trained for manipulation. The simulation comparison isolates two design choices. Comparing our policy with E2E loco-manipulation evaluates the benefit of specialization for manipulation, whereas comparison with the HRL manipulation policy evaluates the effect of optimizing planar manipulation end-to-end rather than through a fixed locomotion policy.

We generate 1000 random environments following the training randomization and evaluate the aforementioned baselines. The average initial distance between the object and the goal is 2.5 $\pm$ 1 \textit{m}. We consider a run to be successful, if the pose error is less than 15 \textit{cm} and $15^\circ$. The episode length is set to 15 \textit{s}. %
We report the following quantities: \textbf{Success Rate} (SR), average \textbf{Episode Time} (ET), \textbf{Fall Over} (FO) rate, and \textbf{Energy Consumption} (EC) for successful runs only expressed as $\sum_t \sum_j |\tau_j \dot q_j| \Delta t$. Comparison results are presented in Table~\ref{tab:baselines}.%

\begin{table*}[t]
\centering
\footnotesize
\caption{Comparison with Baselines. Compared to the HRL baseline, our method improves Success Rate and Episode Time. Compared to the E2E loco-manipulation baseline, our policy improves success rate while retaining low Episode Time.}
\label{tab:baselines}
\begin{tabular}{lcccc}
\toprule
\textbf{Method} & \textbf{Success Rate (\%)} & \textbf{Episode Time (s)} & \textbf{Fall Over (\%)} & \textbf{Energy (J)} \\
\midrule
E2E loco-manipulation & 84.50 & 4.15 $\pm$ 1.46 & 3.5 & 1737.55 $\pm$ 704.01 \\
HRL manipulation                  & 83.50 & 10.80 $\pm$ 4.02 & \textbf{0.3} & 1813.61 $\pm$ 1060.31 \\
\textbf{Ours}         & \textbf{93.30} & \textbf{4.09} $\pm$ \textbf{1.31} & 0.8 & \textbf{1680.80} $\pm$ \textbf{678.12} \\
\bottomrule
\end{tabular}
\end{table*}

We find that an E2E loco-manipulation policy shows similar success rate to that of the HRL manipulation policy, however with significantly smaller (ET) due to the unconstrained behavior discussed in~\ref{sec:HRLvsE2E}. Although this formulation still exceeds the HRL solution in efficiency, the conflicting objectives during training do not enable the model to be optimized properly for manipulation, consequently the model achieves a lower success rate than our method. %
We highlight the difference in efficiency between the HRL policy and ours. Higher success rate, coupled with lower mean and standard deviation of (ET) indicates that our method enables the robot to effectively adapt to different object size and weight, resulting in a more robust manipulation behavior. Nonetheless, the HRL policy emerges with the lowest (FO) rate, only slightly lower than our method, mainly due to the constrained body orientation and gait.

\textit{In summary}, the two design choices are complementary. Combining manipulation-specific training with an end-to-end whole-body architecture yields the highest success rate while retaining low task completion time.%

\subsubsection{Hardware Experiments}
\label{hardware mani}
\begin{figure}[t]
    \centering
    \href{https://youtu.be/Hyvp2y9RCVY}{
        \includegraphics[width=0.75\columnwidth]{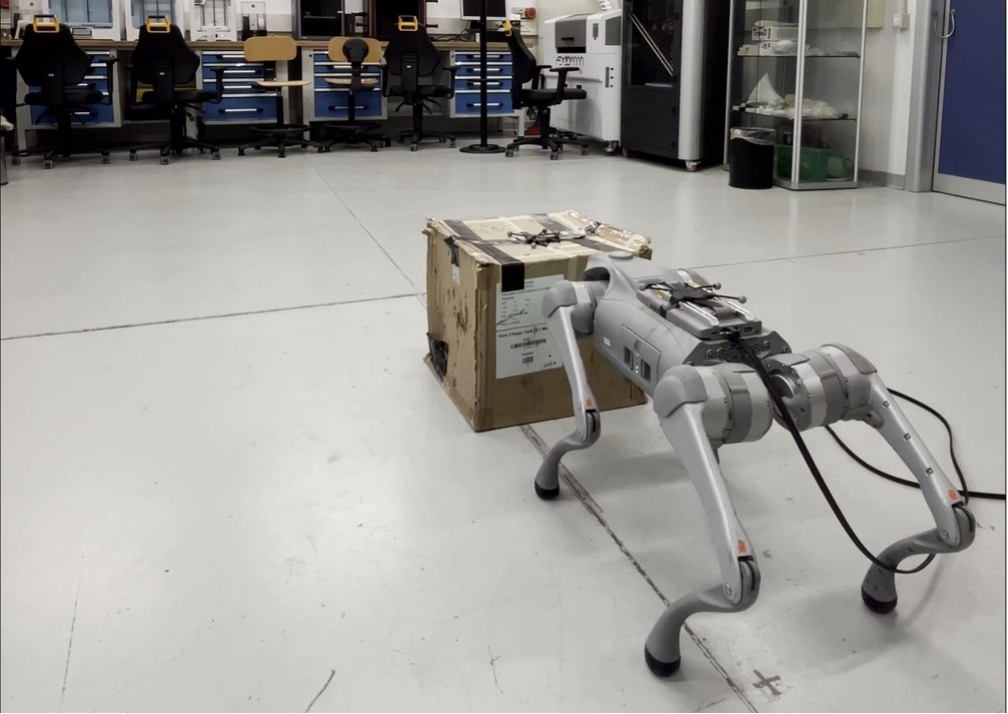}
    }
    \caption{Hardware comparison of the manipulation policies.}
    \label{fig:hardware_video_mani}
\end{figure}

We conduct repeated hardware trials, for both the HRL and the proposed E2E manipulation policies, on three objects with different geometries and masses: \textit{(i)} A cardboard box of edge length 40~[cm] and weight of approx. 2~[kg], \textit{(ii)} A plastic cylinder of height 70~[cm] and diameter of 55 [cm] containing a backpack of 4~[kg] at the bottom, and \textit{(iii)} a plastic rectangular prism containing the box of the Go2 with the battery and charger, weighing approx 8~[kg]. The average initial object-goal distance is approximately $2.0[m]$, $1.4[m]$, and $1.25[m]$; whereas the initial object-goal yaw error is $90^\circ$, $45^\circ$, and $0^\circ$ for the box, cylinder and prism, respectively. The results are presented in Table~\ref{tab:hardware_manipulation}. We provide a video to qualitatively demonstrate the behavioral difference between the runs for both controllers for all three objects in Fig.~\ref{fig:hardware_video_mani}.

The cardboard box is within the training distribution seen in training, whereas the plastic cylinder is outside of the distribution in terms of object height, while the rectangular prism is outside of the distribution in terms of object weight.

\begin{table}[t]

    \centering
    \caption{Hardware evaluation of the manipulation policies.
    Values are reported as mean $\pm$ standard deviation over five trials per object.}
    \label{tab:hardware_manipulation}
    \setlength{\tabcolsep}{3pt}
    \renewcommand{\arraystretch}{1.1}
    \begin{tabular}{llccc}
        \toprule
        \textbf{Object} &
        \textbf{Controller} &
        \textbf{Position Err. [cm]} &
        \textbf{Yaw Err. [deg]} &
        \textbf{Time [s]} \\
        \midrule

        \multirow{2}{*}{Box}
        & HRL & 7.53 $\pm$ 3.12 & 14.20 $\pm$ 7.47 & 4.57 $\pm$ 0.39 \\
        & \textbf{Ours}   & \textbf{ 5.34 $\pm$ 1.53} & \textbf{ 13.12 $\pm$ 5.10} & \textbf{3.08 $\pm$ 0.18} \\
        \midrule

        \multirow{2}{*}{Cylinder}
        & HRL & 13.43 $\pm$ 5.22 & 19.18 $\pm$ 5.92 & 3.06 $\pm$ 0.13 \\
        & \textbf{Ours}   & \textbf{10.37 $\pm$ 2.73} & \textbf{4.99 $\pm$ 2.20} & \textbf{1.88 $\pm$ 0.14} \\
        \midrule

        \multirow{2}{*}{Prism}
        & HRL & 19.44 $\pm$ 9.61 & 12.64 $\pm$ 5.87 & 11.18 $\pm$ 2.09 \\
        & \textbf{Ours}   & \textbf{10.27 $\pm$ 2.68} & \textbf{5.76 $\pm$ 2.31} & \textbf{2.52 $\pm$ 0.44} \\
        \bottomrule
    \end{tabular}
\end{table}

Across all three objects, the proposed policy achieves lower mean pose error and shorter completion time than the HRL baseline. The reduction in completion time is consistent across trials for every object, whereas for the box, which lies within the training distribution, the error distributions of both policies overlap. The gap is most pronounced for the rectangular prism, which represents a shift in object mass relative to the training distribution: position error decreases from 19.44 to 10.27~[cm] and completion time from 11.18 to 2.52~[s]. These results are consistent with the simulation evaluation and indicate that the proposed policy transfers to hardware. %

\subsection{Language-Conditioned Task Planner Evaluation}
\label{Cases}
We evaluate the high-level planner in the procedural geometric planning environment described in~\ref{ss:LLM_planner}, and subsequently evaluate the complete planning and control hierarchy on hardware. The geometric benchmark isolates semantic and geometric decision-making by providing each baseline with the same structured scene representation, while the hardware experiments assess whether the generated skill plans remain executable when coupled with the learned motor skills.

Our planner evaluation aims to address the following questions:
\begin{itemize}
    \item (Q1) How does the finetuned planner compare with a task-specific classical planner and in-context learning baselines?
    \item (Q2) Can the planner compose a novel skill sequence for a shifted task structure?
    \item (Q3) What role does the pretrained backbone play in skill composition and geometric target prediction?
    \item (Q4) Does the resulting planner, trained on synthetic data, transfer to real-world loco-manipulation tasks?
\end{itemize}
We additionally analyze how LoRA rank and action chunking affect both the ID and shifted task families in Appendix~\ref{sec:planner_ablation_model_selection}.

\subsubsection{Evaluation Protocol and Baselines}
\label{evaluation_proto}
\paragraph{Task families} We evaluate the planner across task families that differ in both the required interaction with the environment and the number of high-level skills needed for task completion. We denote navigation and manipulation skills by $\mathrm{(N)}$ and $\mathrm{(M)}$, respectively. The in-distribution (ID) evaluation contains task structures represented in the training distribution. In addition, we evaluate the planner on a shifted task structure whose skill composition is not represented during training. The task families and their skill compositions are summarized in Table~\ref{tab:planner_task_families}.

\paragraph{Baselines} We compare against a classical planner, two in-context learning (ICL) baselines, one using the same pretrained Qwen backbone and one using a frontier model (GPT-5.5), a task-specific Transformer, and a Qwen2.5-architecture model trained from scratch on the same text-based inputs and outputs.

\textbf{Classical planner:} We implement a task-specific classical planner based on a state machine, a visibility graph, Dijkstra's algorithm for geometric path planning, and a sampling-based heuristic for identifying the blocking object to reposition and the relocation target. The classical planner receives the same structured geometric scene information as the LLM planner. Its parameters are tuned jointly across all task families, and a single parameter set is used for all scenes. In addition to the branches covering the training task families, we design a dedicated branch encoding the $\mathrm{M}\rightarrow\mathrm{N}\rightarrow\mathrm{M}$ strategy of the shifted task. The classical planner therefore has explicit knowledge of a task structure that is excluded from our planner’s training data. The choice of this baseline is motivated by the ability of explicitly designing rules and conditions that the plan must meet in order to complete the required task.%

\textbf{ICL baselines:} We compare against two ICL baselines; one using the same pretrained Qwen backbone without any gradient updates, and a frontier model (GPT-5.5). The models receive a fixed prompt describing the available skills, the workspace bounds, the proximity condition for manipulation, and the admissible plan structures, together with one demonstration for each in-distribution task family and for the shifted task family. We additionally include a counter-example for direct navigation in a scene where a free path to the goal is absent. At inference time, the ICL baselines receive the same natural-language task and structured scene states and parameters as the finetuned planner, and generate the complete skill sequence from the initial state, which is then executed open-loop. These baselines evaluate whether the required semantic and geometric planning behavior can be elicited from the same small pretrained Qwen model and a substantially larger model through prompting alone. %
For the evaluation of GPT-5.5 on the shifted task, the description and demonstration of the shifted structure are removed from the prompt, so that, like our planner, it is not informed of it. As for the Qwen ICL baseline, since it fails on the shifted task even with this information, it is reported with the full prompt.

\textbf{Task-specific Transformer:} To assess whether the task structures in our formulation can be learned without pretrained representations, we train a decoder-only Transformer from scratch on the same synthetic dataset. The task objective, robot state, previous skill, and individual scene objects are embedded separately. The model autoregressively predicts the skill, the relevant object through a pointer over the contextualized object representations, and the geometric target $(x, y, \psi)$, which is quantized into bins of 1~[cm] and 1~[$^\circ$] and trained with a cross-entropy loss. As with our planner, the model is trained to predict the remaining action sequence, and only the first action is executed before re-planning. Architecture and training details are provided in Appendix~\ref{sec:planner_pretraining_ablation}.

Additional information on the baselines is provided in Appendix~\ref{sec:sim_results_LLM_A}.

\paragraph{Plan-success criterion} Many scene configurations admit multiple valid solutions, thus we evaluate the generated plan according to task-level feasibility. A plan is considered successful if its execution in the geometric environment completes the requested task without violating the planner's operational constraints. Specifically, (i) navigation and manipulation must remain collision-free with respect to unintended objects and within the workspace boundaries; (ii) manipulation may only be applied to movable objects and only when the robot satisfies a required proximity condition represented by a center-to-center distance of at most 1.2 $[m]$
; and (iii) the resulting sequence of parametrized actions must achieve the task objective. Any violation of these constraints is counted as a planning failure.%

\subsubsection{Simulation Comparison}
\label{sec:Sim_comparison_llm}

We evaluate our planner and the baselines on 200 held-out trajectories spanning the in-distribution task families described in~\ref{evaluation_proto}, and 30 trajectories from the shifted task structure. The results are presented in Table~\ref{tab:baselineresults_LLM}.

\begin{table*}[t]
    \centering
    \caption{Planner baseline results, represented as (\# of successes)/(\# of trajectories) for each task family. Results of our planner and the task-specific Transformer are from a single training run; results of additional runs with different training seeds are reported in Appendix~\ref{sec:seeds}.}
    \label{tab:baselineresults_LLM}
    \setlength{\tabcolsep}{4pt}
    \begin{tabular}{l c c c c c}
        \hline
        \textbf{Task Family}
            & \textbf{Ours}
            & \textbf{ICL (Qwen2.5-1.5B)}
            & \textbf{ICL (GPT-5.5)}
            & \textbf{Transformer (scratch)}
            & \textbf{Classical planner} \\
        \hline
        Direct navigation
            & 38/40
            & \textbf{40/40}$^{\S}$
            & 36/40
            & 35/40
            & \textbf{40/40} \\
        Object manipulation
            & \textbf{10/10}
            & 0/10
            & \textbf{10/10}
            & 3/10
            & \textbf{10/10} \\
        Interactive navigation
            & \textbf{109/120}
            & 0/120
            & 94/120
            & 101/120
            & 86/120 \\
        Long-horizon interactive navigation
            & \textbf{29/30}
            & 0/30
            & 18/30
            & 22/30
            & 22/30 \\
        \hline
        Interactive manipulation (shifted)
            & 12/30
            & 0/30$^{\ddagger}$
            & 2/30
            & 0/30
            & \textbf{20/30}$^{\dagger}$ \\
        \hline
        \textbf{Per-action Latency (ms)}
            & $472.4 \pm 15.5$
            & $953.5 \pm 97.1$
            & $7011 \pm 3400^{\P}$
            & $\mathbf{26.9 \pm 6.2}$
            & $318.4 \pm 598.0^{*}$ \\
        \hline
    \end{tabular}

    \vspace{1mm}
    \begin{minipage}{\textwidth}
        \footnotesize
        $^{\dagger}$The classical planner uses a dedicated branch designed for the shifted task structure; its skill composition is therefore correct by construction.
        $^{\ddagger}$The Qwen ICL prompt describes and demonstrates the shifted task structure, which is excluded from the training data of our planner.
        $^{\S}$The model outputs a single action that mirrors the verb of the task instruction in all scenes; success on direct navigation therefore does not reflect planning.
        $^{\P}$Queried via the OpenAI API; includes network overhead (see Appendix~\ref{sec:sim_results_LLM_A}).
        $^{*}$The classical planner exhibits a highly skewed latency distribution (first quartile: $1.2$ ms; median: $102.1$ ms; third quartile: $276.1$ ms). The larger variance arises primarily in interactive tasks, where the planner samples candidate object relocations and repeatedly recomputes geometric paths.
    \end{minipage}
\end{table*}

\textbf{(Q1) How does the finetuned planner compare with a task-specific classical planner and in-context learning baselines?}

Our planner shows its advantage once successful task completion of the ID task families requires composition of navigation and manipulation skills. Our planner solves 109/120 interactive navigation trajectories, compared with 86/120 for the classical planner. A similar trend is observed for long-horizon interactive navigation, where our planner solves 29/30 trajectories, compared with 22/30 for the classical planner.%

Of the eleven failures of our planner on interactive-navigation tasks, six correspond to incorrectly issuing a direct-navigation skill despite the path being obstructed. Two other failures result from an incorrect geometric target and an incorrect object identifier, respectively. In the remaining three failures, the planner generates a task plan with composition $\mathrm{N}\rightarrow\mathrm{M}\rightarrow\mathrm{N}$. The composition is not inherently faulty, in fact, the planner attempted to align the robot with the object to manipulate. Nevertheless, the manipulation target fails to create a free path, preventing successful completion of the task. The classical planner remains a competitive baseline, but its performance is sensitive to the geometric heuristics used to determine goal reachability and object relocation. Of its 34 interactive-navigation failures, six occur because no suitable relocation target is found, and the remainder because the visibility graph returns a path that is not physically feasible. More conservative obstacle inflation reduces such failures but triggers unnecessary manipulation in direct-navigation scenes, illustrating the trade-off of a single fixed parameterization.

The Qwen ICL baseline outputs a single action mirroring the task instruction in every scene, and therefore solves none of the task families that require combining navigation and manipulation. GPT-5.5 narrows this gap, but our planner matches or exceeds it on all ID task families, most clearly on interactive navigation (109/120 vs.\ 94/120) and long-horizon interactive navigation (29/30 vs.\ 18/30), where its failures are predominantly geometric, at roughly $15\times$ the per-action latency (Appendix~\ref{sec:sim_results_LLM_A}).

\textit{In summary}, the finetuned planner learns the geometric relations underlying interactive navigation without handcrafted rules or large prompts, and exceeds both the classical planner and the prompted frontier model on interactive navigation tasks, solving 109/120 interactive-navigation and 29/30 long-horizon trajectories, consistently across training runs (Appendix~\ref{sec:seeds}).%

\textbf{(Q2) Can the planner compose a novel skill sequence for a shifted task structure?}

We evaluate the planner on the shifted interactive manipulation task, whose task structure ($\mathrm{M}\rightarrow\mathrm{N}\rightarrow\mathrm{M}$) is excluded from the training distribution. The training distribution contains obstacle-clearing behavior through interactive-navigation tasks and object-centric manipulation in scenes where only the target object and the robot are present. The interactive manipulation task requires these behaviors to be recombined into a new sequence: the planner must first manipulate a blocking object, navigate to the specified target object, and subsequently manipulate it to the goal. Our planner completes 12/30 shifted-task trajectories, compared with 20/30 for the classical planner, which solves this task through a branch designed specifically for this structure, and 2/30 for GPT-5.5, whose prompt, like our training data, neither describes nor demonstrates this structure. When the structure is described and demonstrated in its prompt, GPT-5.5 completes 22/30 with the correct composition in all 30 trajectories (Appendix~\ref{sec:sim_results_LLM_A}).

Our planner generates the intended skill composition in 24/30 trajectories (16/30 and 20/30 for the two additional training runs; Appendix~\ref{sec:seeds}). The six failures result from reverting to the familiar object-manipulation sequence seen during training. In the remaining 12 failures, the correct skill sequence is recovered, but the navigation target following manipulation of the blocker does not place the robot sufficiently close to the target object to satisfy the proximity constraint. This behavior is particularly apparent when a large number of objects is present in scene. %

We further probe a longer interactive-manipulation task requiring $\mathrm{N}\rightarrow\mathrm{M}\rightarrow\mathrm{N}\rightarrow\mathrm{M}$, in which the blocker is initially out of reach. For the planner run reported in Table~\ref{tab:baselineresults_LLM}, only 1/30 plans satisfy all constraints, yet the planner generates the intended skill sequence in 22/30 trajectories while selecting the correct objects to manipulate; failures are dominated by geometric targets that violate reachability, do not match the task target, or fail to create a free path. A detailed failure breakdown is provided in Appendix~\ref{sec:LHIN}.

\textit{In summary}, the learned planner can extend the discrete skill composition beyond the structures represented during training and correctly selects the objects to manipulate, but the geometric targets increasingly violate constraints that are only implicitly represented in the training data.

\textbf{(Q3) What role does the pretrained backbone play in skill composition and geometric target prediction?}

To separate the contribution of the planning formulation from that of the pretrained backbone, we compare our planner against the task-specific Transformer. On the shifted task, the Transformer produces the intended $\mathrm{M}\rightarrow\mathrm{N}\rightarrow\mathrm{M}$ composition in all 30 trajectories but completes none (27 and 30 compositions and at most 2/30 completions across training runs in Appendix~\ref{sec:seeds}). Its failures are dominated by geometric targets: after manipulating the blocker, the predicted navigation target directs the robot toward the final task goal rather than to the target object. Although its navigation target resembles that of the interactive-navigation task seen during training, the Transformer does not reduce the shifted task to interactive navigation since it subsequently emits the final manipulation of the correct target object, completing the M$\rightarrow$N$\rightarrow$M composition. A similar pattern appears on the ID families, where the Transformer trails our planner on all four, most notably object manipulation (3/10 vs.\ 10/10), in which it predicts the expected $\mathrm{N}\rightarrow\mathrm{M}$ sequence but places the robot too far from the object to manipulate it. A Qwen2.5-architecture model trained from scratch on the same text-based inputs and outputs performs considerably worse, reaching 63/120 on interactive navigation and composing none of the shifted-task trajectories (Appendix~\ref{sec:scratch_qwen}). %

\textit{In summary}, these results indicate that skill composition does not require pretraining when the input and output representation is structured, whereas pretraining compensates for a generic text interface and additionally yields more reliable geometric targets from limited synthetic data, at the cost of higher inference latency. %

As the models trained from scratch also differ in size from the pretrained backbone, this comparison does not fully isolate the effect of pretraining.

\subsubsection{Hardware experiments}
\textbf{(Q4) Does the resulting planner, trained on synthetic data, transfer to real-world loco-manipulation tasks?}

All hardware experiments use the planner run reported in Table~\ref{tab:baselineresults_LLM}.

\begin{figure}[t]

    \centering

    \href{https://youtu.be/HLqIe3deXKw?si=Nu53_tc0LTuWum4L}{
        \includegraphics[width=0.82\columnwidth]{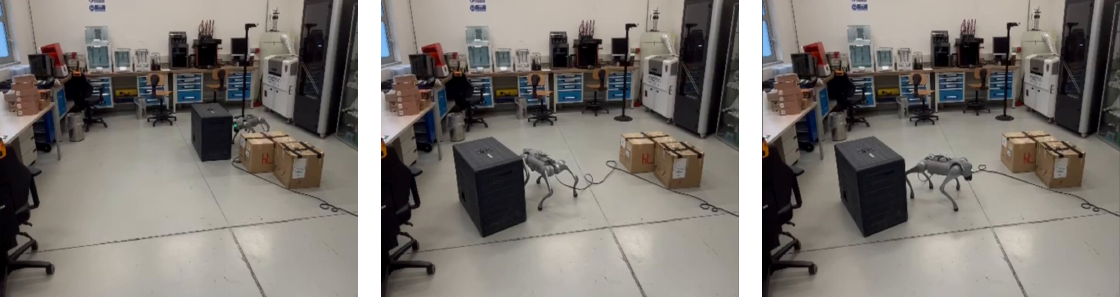}
    }

    {\footnotesize \textbf{(a)} Scene IN-2}

    \vspace{0.5em}

    \href{https://youtu.be/QMH2MtiC1UQ?si=yuf0b7QZQ5aOYLZK}{
        \includegraphics[width=0.82\columnwidth]{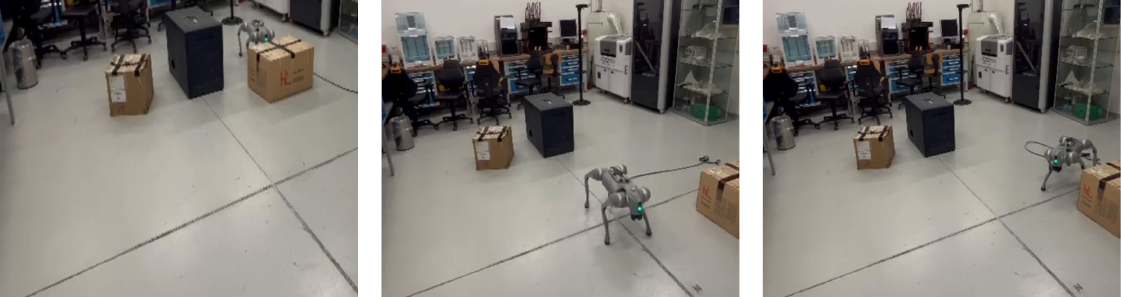}
    }

    {\footnotesize \textbf{(b)} Scene IN-3}

    \vspace{0.5em}

    \caption{Qualitative hardware examples evaluating object selection. In both scenes, the planner must identify and manipulate the appropriate blocking object before navigating toward the goal.}
    \label{fig:object_selection_hardware}
\end{figure}

\begin{figure}[t]

    \centering
    \href{https://youtu.be/bpIoxziqFc0?si=I0GTzRrTaz3M4v0_}{
        \includegraphics[width=1.0\columnwidth]{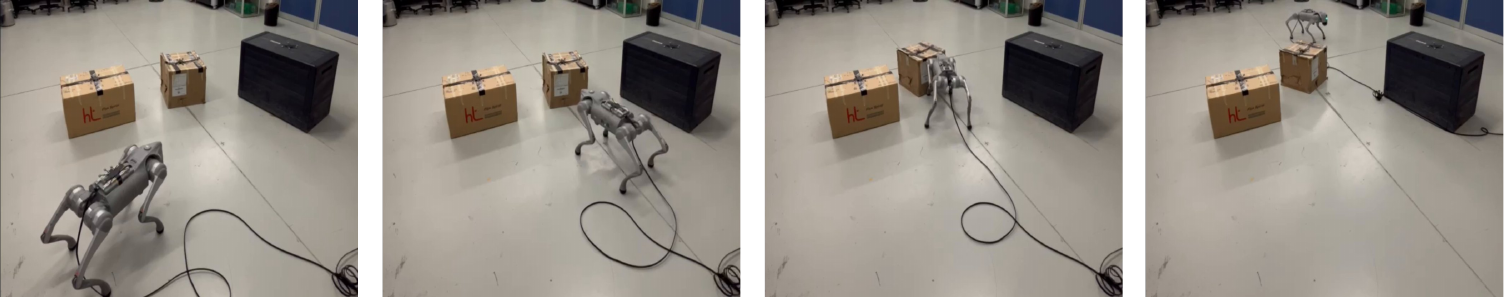}
    }
    \caption{Long horizon task execution for LH-1. The robot appropriately navigates to the only movable object in the scene, manipulates it to create enough space to then navigate to the goal.}
    \label{fig:long_horizon_hardware}
\end{figure}

\begin{figure*}[t]

    \centering

    \href{https://youtu.be/jgPvWF2pvaY?si=L0aW5KpdJ26k7H6y}{
        \includegraphics[width=1.0\columnwidth]{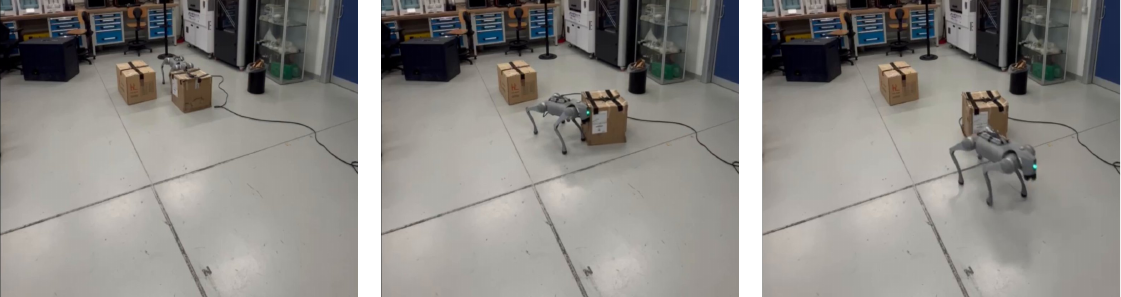}
    }

    {\footnotesize \textbf{(a)} Scene IN-1}

    \vspace{0.5em}

    \href{https://youtu.be/8FpNxKPCOes?si=YWGmnq-jRzsuudY3}{
        \includegraphics[width=1.0\columnwidth]{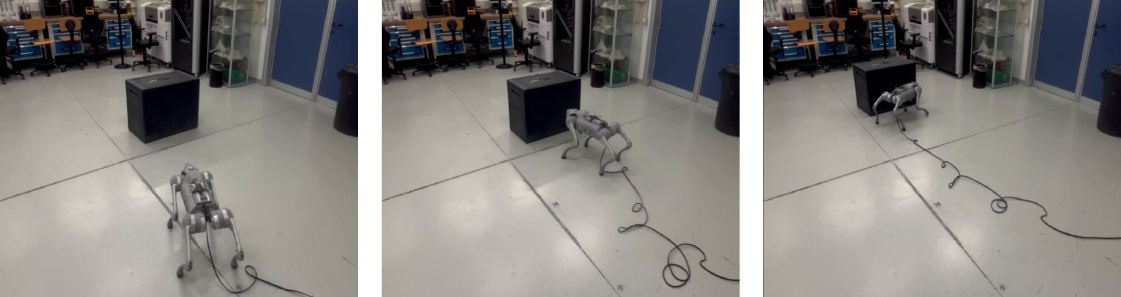}
    }

    {\footnotesize \textbf{(b)} Scene OM-1}

    \vspace{0.5em}

    \href{https://youtu.be/vEV-H5C9CTQ?si=zyLLAY_6LGgpQBbp}{
        \includegraphics[width=1.5\columnwidth]{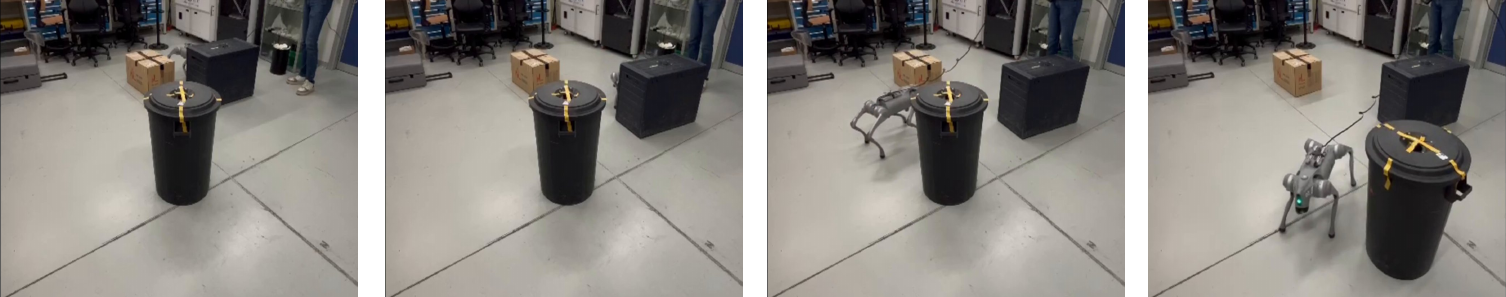}
    }

    {\footnotesize \textbf{(c)} Scene IM-2}

    \vspace{0.5em}
    \caption{Qualitative hardware examples of skill composition. Scene IN-1 requires $\mathrm{M}\rightarrow\mathrm{N}$, whereas Scene OM-1 requires $\mathrm{N}\rightarrow\mathrm{M}$. In Scene IM-2, the planner composes these into the sequence $\mathrm{M}\rightarrow\mathrm{N}\rightarrow\mathrm{M}$, which is excluded from training.%
    }
    \label{fig:Compositional_hardware}
\end{figure*}

We evaluate the complete pipeline on hardware on the following task families: (a) 1 direct navigation case; (b) 1 object manipulation case; (c) 3 interactive navigation tasks, of which 2 qualitatively demonstrate the reasoning capability of the model in choosing the correct object to manipulate; (d) 2 long-horizon interactive navigation tasks; (e) 2 shifted interactive manipulation tasks. Each experiment is repeated 5 times, amounting to a total of 45 runs across 9 scenarios. To highlight at which level the failure occurs, we report the following quantities: (i) correct skill composition, and (ii) end-to-end success. Correct skill composition measures the model's ability in emitting the correct skill sequence. End-to-end success additionally requires correct action parameters and their successful completion on hardware. We note that failure in skill composition is propagated to end-to-end failure. The results are presented in Table~\ref{tab:hardware_planner}. We additionally provide a qualitative analysis of the plans on the scenes discussed above in Figures~\ref{fig:object_selection_hardware},~\ref{fig:long_horizon_hardware}, and~\ref{fig:Compositional_hardware}.

\begin{table*}[t]
\centering
\caption{Hardware evaluation of the complete loco-manipulation pipeline. Each scenario is repeated five times. Correct composition indicates whether the planner selects the appropriate ordered sequence of skills. End-to-end success requires successful completion of the full physical task.}
\label{tab:hardware_planner}
\resizebox{\textwidth}{!}{
\begin{tabular}{lllcccc}
\toprule
\textbf{Family} &
\textbf{Case} &
\textbf{Evaluation focus} &
\textbf{Expected composition} &
\textbf{Trials} &
\textbf{Correct skill composition} &
\textbf{E2E success} \\
\midrule

Direct navigation
& N-1
& Simple Navigation
& $\mathrm{N}$
& 5
& 5/5
& \textbf{5/5} \\

\midrule

Object manipulation
& OM-1
& Simple Manipulation
& $\mathrm{N}\rightarrow\mathrm{M}$
& 5
& 5/5
& \textbf{5/5} \\

\midrule

\multirow{3}{*}{Interactive navigation}
& IN-1
& Task composition
& $\mathrm{M}\rightarrow\mathrm{N}$
& 5
& 4/5
& \textbf{4/5} \\

& IN-2
& Object selection
& $\mathrm{M}\rightarrow\mathrm{N}$
& 5
& 5/5
& \textbf{5/5} \\

& IN-3
& Object selection
& $\mathrm{M}\rightarrow\mathrm{N}$
& 5
& 5/5
& \textbf{5/5} \\

\midrule

\multirow{2}{*}{Long-horizon interactive navigation}
& LH-1
& Long-horizon composition
& $\mathrm{N}\rightarrow\mathrm{M}\rightarrow\mathrm{N}$
& 5
& 5/5
& \textbf{3/5} \\

& LH-2
& Long-horizon composition
& $\mathrm{N}\rightarrow\mathrm{M}\rightarrow\mathrm{N}$
& 5
& 5/5
& \textbf{3/5} \\

\midrule

\multirow{2}{*}{Shifted interactive manipulation}
& IM-1
& Shifted composition
& $\mathrm{M}\rightarrow\mathrm{N}\rightarrow\mathrm{M}$
& 5
& 5/5
& \textbf{3/5} \\

& IM-2
& Shifted composition
& $\mathrm{M}\rightarrow\mathrm{N}\rightarrow\mathrm{M}$
& 5
& 5/5
& \textbf{1/5} \\

\midrule
\multicolumn{4}{r}{\textbf{Completed experiments}}
& \textbf{45}
& \textbf{44/45}
& \textbf{34/45} \\

\bottomrule
\end{tabular}}

\end{table*}

During repeated hardware experiments, manual replacement of the robot and objects introduces small variations in the initial scene configuration. Across the in-distribution task families, the planner nevertheless produces the correct task composition in 34/35 trials. The only compositional error occurs in IN-1, where the planner selects direct navigation to the goal despite the path between the objects being insufficient for collision-free navigation. In this trial, manual resetting of the scene resulted in a wider gap between the objects than in the nominal configuration; however, the gap remained slightly narrower than the robot's width.

At the level of complete physical execution, the pipeline successfully completes 30/35 in-distribution trials. Apart from the skill composition error in IN-1, the remaining failures arise either due to incorrect geometric target selection or low-level execution. In the long horizon case LH-1, one failure results from the manipulation policy not reaching the prescribed terminal accuracy threshold (i.e., the planner was not called for inference at the next decision state), while the other is due to the manipulation target generated by the planner placing the object close to the final target, thus the subsequent navigation stage does not admit a collision-free path to the goal. In LH-2, both failures occur due to the manipulation policy driving the robot outside the room boundary.

We additionally evaluate the planner on two \textit{shifted} interactive manipulation cases requiring the $\mathrm{M}\rightarrow\mathrm{N}\rightarrow\mathrm{M}$ skill composition. The planner predicts the correct composition in 10/10 trials, while the complete pipeline succeeds end-to-end in 4/10 trials. At the skill level, the planner consistently identifies that the blocking object must first be manipulated before navigating toward and manipulating the target object. We further inspect the token probabilities associated with the first manipulation decision and object selection. The model assigns high probability to the manipulation prefix and to the selected blocking object, with an average token probability of $P_{avg}=0.923$ across the aggregated tokens; however, at the final token related to the generated skill name, the probability is split between \texttt{\_to} and \texttt{\_object}. The latter would produce \texttt{manipulate\_object}, which is not part of the defined skill vocabulary. Thus, while the model exhibits some uncertainty in the exact output representation, the probability mass remains associated with manipulation rather than direct navigation to the target object.

The deployment failures in the shifted task IM-1 were associated with navigation targets that placed the robot at object-relative distances outside those observed during training of the manipulation policy, resulting in inconsistent low-level execution. In IM-2, the errors were instead related to the yaw command of the navigation target. These failures can be interpreted from both the planner and low-level policy perspectives. On the low-level side, the manipulation policy learns to manipulate the object predominantly with its left side, irrespective of the relative pose between the robot and the object. Thus, when a navigation action repositions the robot with the object to its right, the resulting state lies outside of the manipulation policy's training distribution, leading to inconsistent execution. At the higher level, the underlying issue is that the planner generates geometric targets without knowledge of the execution capabilities of the parametrized actions. Consequently, correcting such failures only at the low level would require post-hoc adaptation or retraining of the manipulation policy to accommodate planner-generated states. A tighter coupling between high-level target generation and low-level policy dynamics would instead allow the planner to avoid issuing commands that are difficult for the downstream policies to execute reliably.

\textit{In summary}, the planner trained on synthetic data transfers to hardware with the correct skill composition in 44/45 trials, while end-to-end success is limited by the decoupling between planner-generated targets and low-level execution, particularly for long-horizon and shifted tasks.

Finally, the examples in Figures~\ref{fig:object_selection_hardware},~\ref{fig:long_horizon_hardware}, and~\ref{fig:Compositional_hardware} illustrate the different task-level decisions evaluated above. In Figure~\ref{fig:object_selection_hardware}, the two objects near the robot are movable obstacles, whereas the third is immovable. Between IN-2 and IN-3, the immovable obstacle is placed behind a different movable object, changing which object must be displaced to create a feasible route to the goal. The planner correctly changes its object selection according to this geometric relation. The LH-1 example in Figure~\ref{fig:long_horizon_hardware} illustrates the longer-horizon $\mathrm{N}\rightarrow\mathrm{M}\rightarrow\mathrm{N}$ composition, where the robot must first navigate to the blocker, manipulate it to clear the path, and subsequently navigate to the final goal. Finally, Fig.~\ref{fig:Compositional_hardware} contrasts task structures represented during training with a successful execution of the shifted IM-2 case, in which previously learned navigation and manipulation behaviors are recombined into the novel $\mathrm{M}\rightarrow\mathrm{N}\rightarrow\mathrm{M}$ sequence.

\vspace{-0.5em}
\subsection{Limitations and Future Work}
Although the proposed framework offers a quick and deployable recipe for finetuning LLMs for loco-manipulation tasks, one limitation is that the planner is only inferenced at discrete transition steps. Future work should investigate more frequent inference, enabling better plan adaptation under changes in the environment. Additionally, our evaluation is restricted to a single room configuration; future work should extend the framework across different room layouts. Another limitation of the framework is that the planner has no explicit model of the execution feasibility of its parameterized skills, leading to end-to-end failures on hardware. Finally, a natural extension of this project is to rely on local information (i.e. perception) rather than assuming that the pose of the entities in the scene is available.%

\section{Conclusion}
\label{ss:conclusion}

In this work, we proposed a hierarchical framework for quadruped planar loco-manipulation in cluttered environments. The high-level planner, a language-conditioned task planner finetuned with LoRA on synthetic data, jointly selects the skill, the relevant object, and a geometric target, and learns the geometric relations underlying interactive navigation without handcrafted rules or large prompts. Comparisons with models trained from scratch indicate that skill composition, including into a task structure excluded from training, does not require pretraining when the representation is structured, whereas pretraining yields more reliable geometric targets, which remain the main limitation of full-task success. At the low level, the proposed end-to-end whole-body manipulation policy adapts its behavior to a wide range of object properties and exhibits an efficient planar pushing behavior, improving success and reducing task-completion time by more than $60\%$ compared with the commonly used hierarchical approach. On hardware, the complete framework produces the correct skill composition in 44/45 trials and succeeds end-to-end in 34/45. %

\bibliographystyle{IEEEtran}
\bibliography{references}
\clearpage
\input{appendix_body}

\end{document}

%% file: appendix_body.tex
\appendices
\section{Notation}

\textbf{General:} Unless otherwise noted, boldface denotes \textbf{vectors}, and plain symbols denote scalars. We use the following reference frames: $W$ (world), $R$ (robot/base), $O$ (object), and $G$ (goal). For any quantity $\square$, ${}^{F}\!\square_{AB}$ denotes the quantity of $B$ relative to $A$, expressed in frame $F$; e.g., the position of the goal relative to the object, expressed in the robot frame is denoted as ${}^{R}\!\mathbf{p}_{OG}$. The unit normalized vector is $\hat{\mathbf{a}}$, and $\lVert\cdot\rVert$ denotes the Euclidean ($\ell_2$) norm. Yaw angles are denoted by $\psi\in(-\pi,\pi]$ (a coordinate on $\mathrm{SO}(2)$). The actor observations are denoted by \textit{$o_{s}$} where $s=[n,m]$ refers to navigation and manipulation skills, respectively.

\textbf{Robot-related quantities:} The joint configuration is $\mathbf{q}\in\mathbb{R}^{12}$ with the default (nominal) configuration $\mathbf{q}_0$ and offsets $\Delta \mathbf{q}= \mathbf{q} - \mathbf{q}_0$; joint velocities and torques are $\dot{\mathbf{q}}$, and $\boldsymbol{\tau}$, respectively. The base (body) pose is given by ${}^{W}\!\mathbf{p}_{WR} \in \mathbb{R}^3$ (position of $R$ in $W$), and yaw $\psi_R$ (rotation about the $W$ $z$-axis). The base twist comprises linear and angular velocities expressed in $R$, denoted by $\mathbf{v}_R \in \mathbb{R}^3$ and $\boldsymbol{\omega}_R \in \mathbb{R}^3$. $v_{R,i}$ denotes the \textit{i-th} component of the velocity vector of the robot $\mathbf{v}_{R}$.

\textbf{Object-related quantities:} the object pose is ${}^{W}\!\mathbf{p}_{O}\in\mathbb{R}^3$ (position of $O$ in $W$) with yaw $\psi_O$; the object twist is $\mathbf{v}_O\in\mathbb{R}^3$ and $\boldsymbol{\omega}_O\in\mathbb{R}^3$ (linear and angular velocities of $O$). The goal pose is ${}^{W}\!\mathbf{p}_{G}\in\mathbb{R}^3$ with desired yaw $\psi_G$. %

\section{Training details}
\subsection{Training and platform details}
\label{sec:TrainingA}
Policy training was performed on a workstation equipped with an Intel\textsuperscript{\textregistered} Core\textsuperscript{\texttrademark} Ultra 9 275HX CPU (24 cores) and an NVIDIA RTX 5080 GPU. The manipulation policy $\pi_m$ was trained for 8000 iterations, while the high-level navigation planner $\pi_n$ was trained for 6000 iterations. Manipulation and locomotion policies were trained using 4096 and 1024 parallel environments, respectively, with 24 environment steps per iteration. Training required approximately 3 hours for the manipulation policy and 4 hours for the locomotion policy.

LLM fine-tuning was performed on two NVIDIA RTX 5090 GPUs and required approximately 45 minutes. During deployment, action inference runs at 1 Hz on an RTX 5080 GPU.
\subsection{Navigation policy}
\label{sec:navigationA}
The inputs to the navigation policy are proprioception information $\textit{$o_{n,t}$} \in \mathbb{R}^{27}$ (joint positions w.r.t. default configuration, joint velocities and base angular velocity), previous actions $\mathbf{a_{t-1}\in \mathbb{R}^{3}}$, pose commands $\mathbf{c_t}$, and depth data $\mathbf{h_t}\in \mathbb{R}^{7}$ coming from the on-board LiDAR. $\mathbf{h_t}$ represents the maximum depth from the sensor to the ray hit location for seven discrete ray angles $\in [-\pi/2,\pi/2]$. This choice of environment sensing is due to the built-in L1 sensor of the Unitree Go2, which has a non-repetitive scan pattern and produces sparser depth readings than other commonly used sensors.

The terrain consists of a varying number of randomly generated rectangular obstacles with varying widths and heights, as done in~\cite{rudin2022advancedskillslearninglocomotion}, positioned in a square of dimension $4\times4$ [$\text{m}^2$] centered at the origin of the environment. An area of dimension $1\times1$ [$\text{m}^2$] is left clear at the center of the environment, and serves as the robot's spawning platform. Position commands are sampled in polar coordinates outside of the obstacle area, while goal and robot yaw are sampled to cover the entire space $\in [-\pi,\pi]$.

\subsection{Manipulation policy}
\label{sec:manipulationA}
The complete list of rewards is provided in Table~\ref{tab:manipulation_rewards}. PPO and network hyperparameters are provided in Table~\ref{tab:rl_training_hyperparameters}.
\begin{table}[t]
    \centering
    \caption{Reward terms used for training the manipulation policy.}
    \label{tab:manipulation_rewards}
    \renewcommand{\arraystretch}{1.15}
    \begin{tabular}{lll}
        \toprule
        \textbf{Reward term} & \textbf{Description} & \textbf{Weight} \\
        \midrule

        $rew_{\mathrm{kp}}$
        & Object--goal keypoint tracking
        & $20.0$ \\

        $rew_v$
        & Object velocity tracking
        & $15.0$ \\

        $rew_{v_{R,x}}$
        & Positive robot forward velocity
        & $0.5$ \\

        $\lVert \ddot{\mathbf{q}} \rVert^2$
        & Joint acceleration penalty
        & $-2.5*10^{-7}$ \\

        $\lVert \boldsymbol{\tau} \rVert^2$
        & Joint torque penalty
        & $-2*10^{-4}$ \\

        $\lVert \Delta \mathbf{a}_{t-1,t} \rVert^2$
        & Action-rate penalty
        & $-0.4$ \\

        $\lVert \boldsymbol{\omega}_{O,xy} \rVert^2$
        & Object roll/pitch velocity penalty
        & $-2.0$ \\

        $\lVert \boldsymbol{\omega}_{R,xy} \rVert^2$
        & Robot roll/pitch velocity penalty
        & $-0.2$ \\

        $\lVert \mathbf{a}_{\mathrm{feet}} \rVert^2$
        & Feet acceleration penalty
        & $-8*10^{-3}$ \\

        $\lVert \mathbf{q} - \mathbf{q}_0 \rVert^2$
        & Joint deviation from default configuration
        & $-0.5$ \\

        \bottomrule
    \end{tabular}
\end{table}

The total reward $r$ follows,
\begin{equation}
r = \sum_{i=1}^{N} w_i\, rew_i
\end{equation}
where $rew_i$ denotes each reward term $i$, and $w_i$ its associated weight.

We use an action scale of $a_{scale}=0.25$, and PD gains of $K_p=25$ and $K_d=0.5$. We do not normalize the observations before feeding them to the policy.
\begin{table}[t]
\centering
\caption{PPO and network hyperparameters used for training the low-level policies with RSL-RL.}
\label{tab:rl_training_hyperparameters}
\begin{tabular}{lc}
\hline
\textbf{Parameter} & \textbf{Value} \\
\hline
Rollout steps per update & 24 \\
Action clipping & 5.0 \\
Actor hidden layers & [512, 256, 128] \\
Critic hidden layers & [512, 256, 128] \\
Activation function & ELU \\
Initial action noise std. & 1.0 \\
Value loss coefficient & 1.0 \\
Clipped value loss & True \\
PPO clipping parameter & 0.2 \\
Entropy coefficient & 0.005 \\
Learning epochs per update & 5 \\
Mini-batches per update & 4 \\
Learning rate & $1.0\times10^{-3}$ \\
Learning-rate schedule & Adaptive \\
Discount factor $\gamma$ & 0.99 \\
GAE parameter $\lambda$ & 0.95 \\
Desired KL divergence & 0.01 \\
Maximum gradient norm & 1.0 \\
\hline
\end{tabular}
\end{table}
\section{HRL vs E2E loco-manipulation policies}
\label{sec:HRLvsE2EA}
\begin{figure}[t]
    \centering

    \subfloat[Manipulation reward for the three E2E loco-manipulation
    policies over training steps.]{
        \includegraphics[width=0.48\linewidth]
        {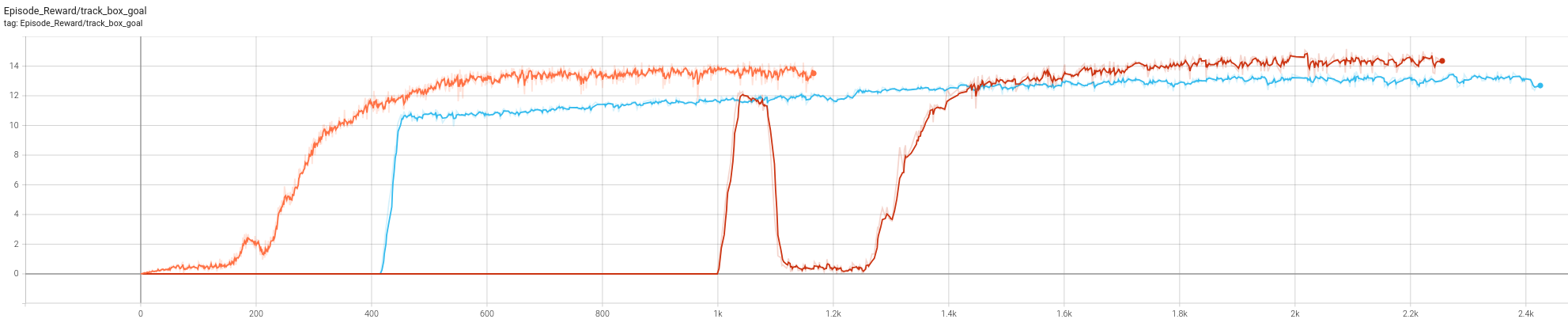}
    }
    \hfill
    \subfloat[Action standard deviation of the three E2E
    loco-manipulation policies over training steps.]{
        \includegraphics[width=0.48\linewidth]
        {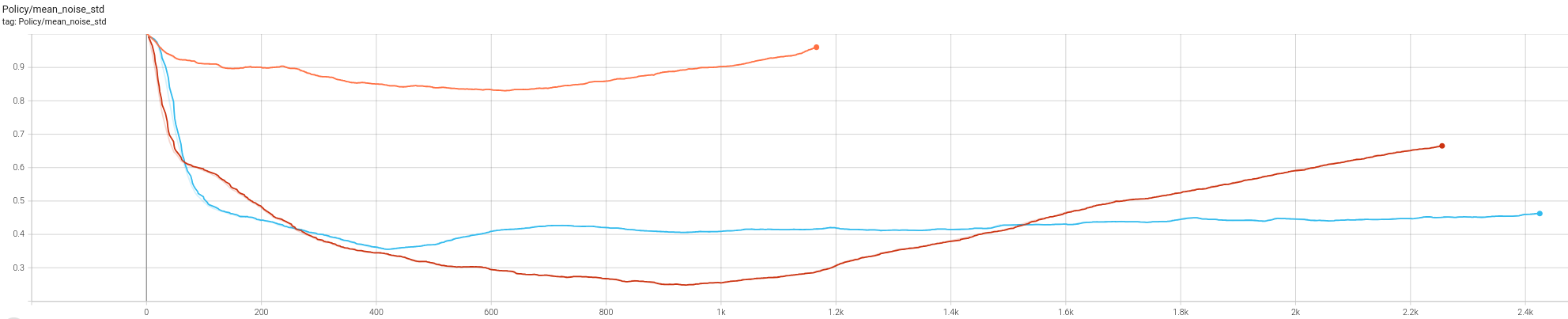}
    }

    \caption{End-to-end loco-manipulation policy learning dynamics.
    Manipulation reward was activated at iteration 0 for the orange
    curve, iteration 400 for the cyan curve, and iteration 1000 for
    the red curve.}
    \label{fig:details_E2E_locomani}
\end{figure}

As discussed in~\ref{sec:HRLvsE2E}, the HRL loco-manipulation policy was trained following the pipeline of~\cite{jeon2023learningwholebodymanipulationquadrupedal}, excluding the adaptation module. Additionally, the low-level velocity tracking policy is slightly simpler than the one implemented (based on\cite{hwangbodeformable}) and does not require any state estimation. Although using the low-level policy from~\cite{hwangbodeformable} could lead to a slightly more performant loco-manipulation policy, it does not solve the imposed constraint on the policy's behavior. We found that training the E2E loco-manipulation policy is trickier than training the HRL policy, similar to what was discussed in~\cite{jeon2023learningwholebodymanipulationquadrupedal}; however, the following method worked consistently well: We first reward the policy for approaching the object with a heading inline with the target vector ${}^{W}\!\mathbf{p}_{OG}$, this stage can be seen as a "warm start" to the policy as it learns a "base" stable behavior. When the robot has learned to approach the object, it will collide with it inducing a motion towards the goal. At this stage, the reward for reposing the object is activated. The initial learned behavior of colliding with the object offers a strong signal for the manipulation phase. The number of training iterations for the first phase is important as activating the manipulation reward too late will cause big variations in policy update. On the other hand, activating it too early and the robot doesn't learn as well due to the conflicting objectives, as demonstrated in figure~\ref{fig:details_E2E_locomani}.

Both policies are trained for the same environment configuration, and randomization. Reward tuning was done separately to ensure a good policy checkpoint has been reached.
\begin{figure}[t]
\centering

\subfloat[\label{fig:run1}]{%
{
\includegraphics[width=0.48\linewidth]{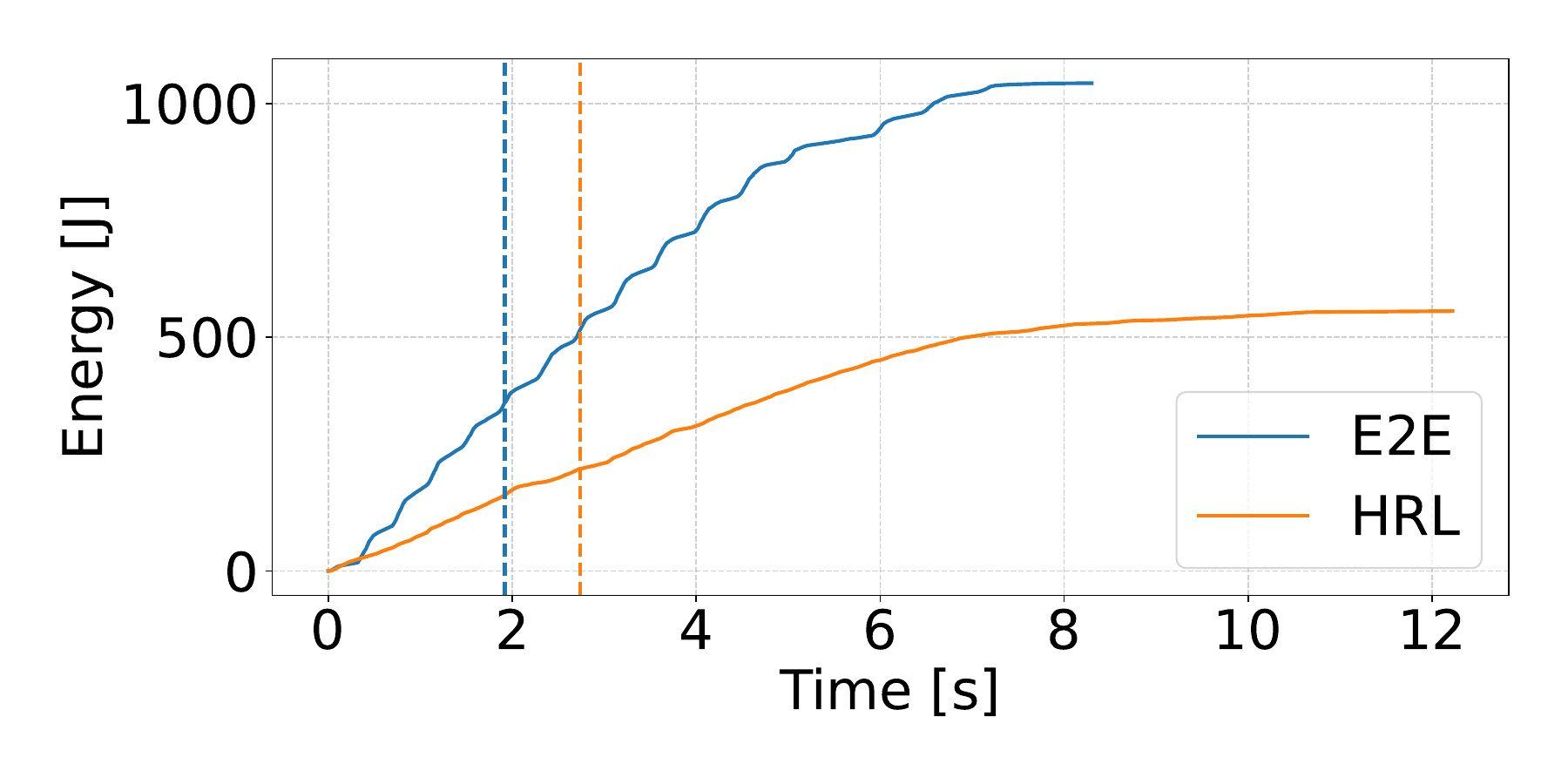}}}
\hfill
\subfloat[\label{fig:run2}]{%
\includegraphics[width=0.48\linewidth]{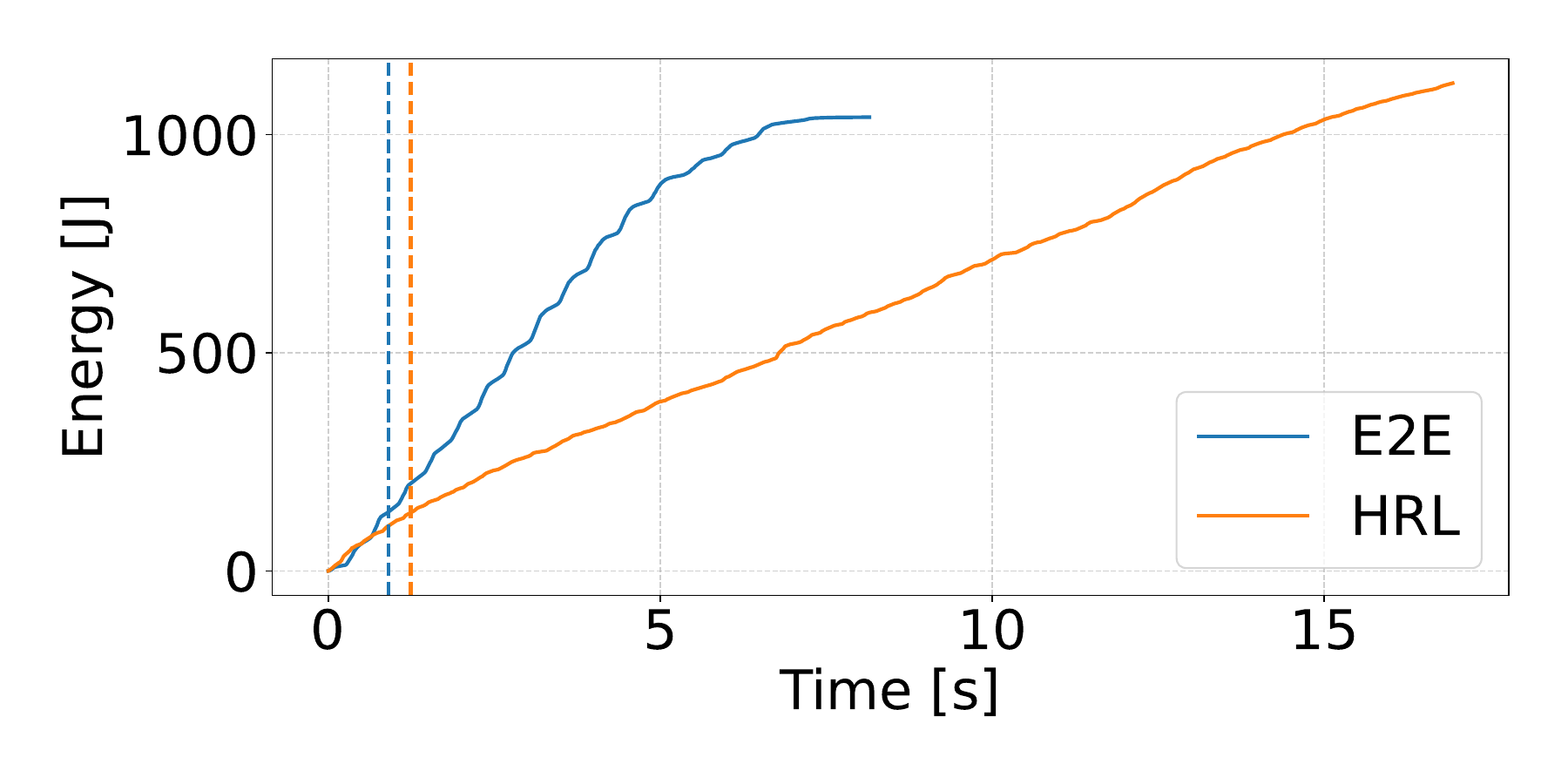}}

\caption{Comparison of Energy Consumption between HRL and E2E Loco-manipulation policies. \textbf{(a) represents the case in which the object distant from the robot}, and \textbf{(b) represents the case in which the object close to the robot}. Blue and Orange \textbf{dashed} lines represent the first instant in which the robot is in contact with the object for E2E and HRL, respectively. In these two runs, the object is a cube with an edge length of 40 [cm], weight and CoM position are inside the training distribution. We provide a video showing the behavioral difference between both methods for the first case.}
\label{fig:energy_comparison}

\end{figure}

Figure~\ref{fig:energy_comparison} shows the energy consumption for both methods for the following two runs: (\textit{a}) The object is distant from the robot, (\textit{b}) the object is close to the robot. In both cases, we see that the E2E locomanipulation policy is significantly faster than the HRL approach in reposing the object to the goal. However, it is only more energy efficient for case \textit{b}. Table~\ref{tab:energy_phases} shows the per-phase energy consumption for each case. We note that the HRL policy consumes less energy in the manipulation phase of the first run, even with a 50$\%$ longer episode time. The E2E approach is rather more energetically efficient for cases where the robot is close to the object, and more notably for a long manipulation phase. We note that the E2E policy shown in this experiment was trained without an explicit penalty on foot accelerations. Incorporating this penalty reduces impact forces on hardware, resulting in lower impact noise and reduced kinetic energy dissipation.

\begin{table}[t]
\centering
\caption{Energy in [J] for locomotion and manipulation phases, for both cases.}
\label{tab:energy_phases}
\begin{tabular}{l c c c c}
\hline
 & \multicolumn{2}{c}{\textbf{Run 1}} & \multicolumn{2}{c}{\textbf{Run 2}} \\
\textbf{Method} & Locomotion  & Manipulation  & Locomotion & Manipulation \\
\hline
E2E & 359.56 & 684.28 & 136.7 & 903.3 \\
HRL & 218.63 & 337.42 & 133.9 & 984.0\\
\hline
\end{tabular}
\end{table}

We note that the reason behind the lower completion time of our method, or end-to-end approach in general, is not due to a higher base velocity. This was shown in section~\ref{sec:HRLvsE2E}, where we compare both loco-manipulation approaches and figure~\ref{fig:body_orientation_comparison} show that both policies achieve the locomotion phase in approximately similar durations. Thus, demonstrating that the end-to-end approach is not \textit{faster} in terms of robot base velocity, it is only \textit{better suited} for manipulation.

\section{Manipulation policy ablation}
\label{sec:Ablationmani}
The ablation study was carried out on the same environment configuration used for the comparison in Section~\ref{ss:Baselines_mani}. Results are presented in Table~\ref{tab:ablation}. In the case where we omitted the positive velocity reward, we observe that (FO) increases by an order of magnitude when compared to our solution, and success rate drops to $20.7\%$. The average error for (TO) runs is 0.28 [m] and $40^\circ$, indicating that the robot is falling over around the end of the episode. Upon closer inspection, the cases related to \textbf{Fall Over} were mostly due to the robot flipping over. Without this reward, the robot chooses to manipulate the object with its lateral side. The problem arises once the robot has to detach from the object. Once no opposing force is present to stabilize the configuration of the robot, it either \textit{(i)} falls over due to the CoM being outside the support polygon, or \textit{(ii)} collides with the object, pushing it away from the goal, and then re-stabilizes itself. Left with little time to re-adjust the pose of the object, $73.3\%$ of the runs end in \textbf{Time Out} (TO). Our method leads to significantly higher success rate since, due to the geometry of the robot, the range of stable pitch angles is bigger than the range of stable roll angles. Also, within the successful runs, we notice that our method emerges with a lower (ET). The robot can displace the object at higher velocities, in part due to the increased stability and the alignment of the calf and thigh motors with the direction of motion of the robot. Instead, when pushing to the side, the motors on the hip joints constitute the main movers. By carefully modulating this reward, we are also able to emerge with a dual effect, where the robot pushes with a forward heading, until it comes close to the goal position, where it rotates the object from the side while having a more stable configuration.

Removing $r_v$ does not affect the success rate drastically, however we notice a ten-fold increase in (FO) rate when compared to ours, mainly attributed to the object falling over. Excluding our $r_v$ reward leads to a faster, more aggressive behavior towards the end of the episode, also reflected by the lower (ET).

\begin{table*}[t]
\centering
\footnotesize
\caption{Ablation Results.}
\label{tab:ablation}
\begin{tabular}{lcccc}
\toprule
\textbf{Method} & \textbf{Success Rate (\%)} & \textbf{Episode Time (s)} & \textbf{Fall Over (\%)} & \textbf{Energy (J)} \\
\midrule
w/o Positive Velocity & 20.70 & 5.86 $\pm$ 1.99 & 6.0 & 2446.19 $\pm$ 687.44 \\
w/o $r_v$             & 87.30 & \textbf{3.77} $\pm$ \textbf{1.09} & 7.5 & 2028.28 $\pm$ 789.67 \\
\textbf{Ours}         & \textbf{93.30} & 4.09 $\pm$ 1.31 & 0.8 & \textbf{1680.80} $\pm$ \textbf{678.12} \\
\bottomrule
\end{tabular}
\end{table*}

\section{LLM Task Planner}
\label{sec:LLMA}
\subsection{Data generation and augmentation}
\label{sec:DatagenerationA}
The cases are ordered by increasing difficulty, where difficulty is defined as a function of (i) the number of required skill switches and (ii) the degree of spatial reasoning necessary to resolve the task objective. This structured decomposition induces a curriculum over task complexity and simplifies scalable data generation and augmentation. The proposed case progression exposes the model to increasingly complex spatial configurations, transitioning from direct goal reaching to scenarios that require explicit reasoning over object placement and skill switching. At this stage, we note that updates are purely deterministic (e.g., navigation sets the robot to the target pose; manipulation places the object at the desired pose). Trajectory generation, therefore, operates at a symbolic, task-level abstraction.

Within Case 2, task difficulty is further modulated by controlling object count and spatial arrangement. Simpler instances include a single object or multiple objects that are well-separated from the robot’s path. More challenging instances involve multiple objects arranged to create narrow yet navigable corridors. This structured variation discourages unnecessary manipulation by exposing the model to a broad distribution of non-blocking configurations, thereby reinforcing correct skill selection when navigation suffices.
We additionally introduce \textbf{Case 2.5}, which corresponds to object-centric commands (e.g., manipulating object "A" to a specified pose). This case isolates the structure of manipulation instructions and their associated scene dependencies (analogous to Case 1 for navigation).

Data generation and augmentation are performed in a case-dependent manner to ensure controlled coverage of increasing task complexity, imposing a deliberate bias towards more challenging scenarios. The following describes in detail data generation and initial augmentation.

\textbf{Easy Cases (Case 1 and simple Case 2):}
For low-complexity scenarios, a small set of base environments is manually constructed to define representative spatial layouts. These base templates are then augmented by sampling new robot poses, goal poses, and (when present) object poses from predefined spatial distributions. The sampling ranges are chosen to preserve the defining characteristics of each case (e.g., maintaining non-blocking configurations in Case 2).

\textbf{Medium-Complexity Cases (Case 2 – Narrow Corridors):}
For more challenging navigation-only scenarios, we manually design a diverse set of narrow but traversable passage configurations. These templates are augmented using the same pose-randomization strategy, but under a different spatial distribution that preserves corridor geometry while varying robot, goal, and object placements. This ensures variability while maintaining the requirement that navigation remains feasible without manipulation.

\textbf{Hard Cases (Case 3 – Obstructed Scenes):}
For obstruction scenarios requiring skill switching (i.e., Interactive Navigation tasks), manual generation plays a larger role to ensure broad semantic coverage. We vary both movable and immovable blocking objects, as this directly affects the required action sequence. Augmentation is again performed via controlled pose randomization, using distributions that preserve the blocking constraint. Ensuring wide coverage in this case is critical; insufficient variation would fail to expose the model to the diversity of manipulation–navigation transitions necessary for robust planning.

Additionally, we apply the following structure-preserving augmentations:
\begin{itemize}
    \item \textbf{Object ID and type randomization:} Object identifiers are uniformly sampled within a fixed range (up to six objects per scene) to prevent overfitting to index-based correlations. We also randomize the type of the object (circle or rectangle while keeping the surface area equal).
    \item \textbf{Transition-level state randomization:} Since the transitions in our synthetic dataset are purely deterministic (which is not representative of the real world), we insert a uniform noise to the initial robot and object pose (ensuring no semantic change has occurred), and to the state transitions caused by the actions. This is different from the initial augmentation done where the goal was to cover as much of the state space for each entity.%
    \item \textbf{Symmetry transformations:} Scenes are mirrored across spatial quadrants, yielding up to four equivalent geometric variants per configuration.
    \item \textbf{Irrelevant object injection:} One or two irrelevant objects are inserted at non-interfering positions to increase robustness to irrelevant objects without altering the ground-truth action sequence.
\end{itemize}

Once the augmented trajectories are obtained, they undergo a feasibility verification step to ensure collision-free execution and consistency with the intended task constraints. Any invalid configurations are discarded. Approximately $85\%$ of the generated trajectories were automatically generated.

\subsection{LLM finetuning}
\label{sec:LLMfinetuningA}

We choose to finetune the model with LoRA , since~\cite{kim2024openvlaopensourcevisionlanguageactionmodel} report minor improvements w.r.t full finetuning at a fraction of the cost.
While in our setting the predicted action chunks are considerably shorter compared to VLA models for robotic manipulators~\cite{zhao2023learningfinegrainedbimanualmanipulation,liu2025bidirectionaldecodingimprovingaction,kim2025finetuningvisionlanguageactionmodelsoptimizing,black2025trainingtimeactionconditioningefficient,wei2026psi0openfoundationmodel} due to our task-level planner design, we find that predicting the complete action sequence enables the model to compose the behaviors learned during training to form novel skill sequences.

Although the aforementioned studies have shown that LLM planners, through AC (and its more intricate variants such as RTC~\cite{black2025trainingtimeactionconditioningefficient}), can output consistent long sequences of actions, we opt to train the model on discrete skill-transition actions. This design choice presents a natural mitigation strategy for compounding errors and is motivated by the inherent complexity in the deployment of legged robots, and the task we are solving. %

Specifically, the LLM planner would have to output sequences of geometric targets that are aligned with the low-level capabilities of the robot. Such a geometric output is outside the current capabilities of the model since it was finetuned on synthetic data. This also motivates the choice behind inferring the planner once the previous skill is done, and the robot has acquired a stable configuration.
We would like to note that the current model could (in theory) be inferenced continuously rather than waiting for previous actions to be completed; however, the LLM does not have access to robot proprioception information that would allow it to infer the motion state of the robot (stationary or in motion). The problem arises when the robot has a non-zero base velocity. To enable continuous inference, the LLM has to first have access to such information and must be able to output \textit{idle} actions. For example if the LLM gives a conservative manipulation target (which is representative of the expert data we generated) to make enough space to navigate, and we continuously inference the model, the skill (and consequently its target) might change given that the model reasons that now the robot has enough navigation space. This would lead to a big Sim2Real gap, since (1) the LiDAR data would effectively identify the contact as a collision, and (2) the robot has to change its behavior from manipulation to locomotion in a very short time span. Ideally, the LLM would have access to proprioception information onboard the robot, and output an \textit{idle} target that is close to the current object position. This would ensure that the transition between the different skills is smooth and does not lead the robot to fall over. Unfortunately, the current planner is unable to infer such actions. We leave this step for future work.

Table~\ref{tab:llm_training_details} summarizes the hyperparameters used for training the LLM. We note that the batch size is smaller than in prior work, leading to increased gradient noise and partial GPU under-utilization. However, with appropriate learning rate tuning, the model converges reliably. This batch size was selected as it yields the best empirical performance among the tested values $[8,16,32,64,128]$. A batch size of 128 represents the largest batch size achievable on our hardware without encountering out-of-memory errors. We note that we split the trajectories at the trajectory level to avoid leakage between subsequent decision states of a trajectory. Held-out ID trajectories may share the same task-case representation as training trajectories but contain independently perturbed robot, goal, and object configurations.%

\begin{table}[t]
\centering
\caption{LLM fine-tuning details.}
\label{tab:llm_training_details}
\begin{tabular}{l c}
\hline
\textbf{Parameter} & \textbf{Value} \\
\hline
Dataset size & ~16K I/O pairs (8096 trajectories) \\
Validation dataset size & 5$\%$ of total dataset \\
LoRA rank & 128 \\
LoRA alpha & 256 \\
LoRA dropout & 0.05 \\
Learning rate & 2e-5 \\
LR schedule & Cosine with warmup \\
Warmup ratio & 0.02 \\
Batch size & 16 \\
\hline
\end{tabular}
\end{table}

\subsection{Inference Procedure}

During inference, the language-conditioned planner receives the current scene state from the OptiTrack system and predicts the complete remaining sequence of actions. However, only the first action is executed by sending its target to the corresponding low-level policy (navigation or manipulation). An action is considered complete once the relevant controlled entity reaches a predefined positional accuracy threshold of 15~[cm] and 15 $\circ$ with respect to the commanded target and the robot base velocity, estimated from a temporal sequence of robot-pose observations, falls below 0.01~m/s. At this point, the planner is queried again using the updated scene state, with the previously executed skill additionally provided as input.

If the active low-level policy fails to satisfy the completion criterion, the planner is not queried for the next action and does not intervene by issuing an alternative command. The corresponding run is therefore counted as an end-to-end failure.

\section{Planner Evaluation Details}
\label{sec:sim_results_LLM_A}

Throughout, per-action latency is the query time divided by the number of generated actions.
\paragraph{Classical planner}

We design separate branches for each task, i.e. navigating to a goal pose or manipulating an object to a goal pose. When the navigation goal is unreachable, the planner considers movable objects as potential blockers, samples candidate relocation targets, and counter-factually updates the scene to determine whether each manipulation restores a feasible path. Manipulation is permitted only when the robot is sufficiently close to the selected object; otherwise, the planner first generates a navigation target that aligns the robot with the intended object-motion direction. An analogous branching strategy is used for object-manipulation tasks when access to the target object is obstructed. Planner parameters are tuned jointly across all task families, and a single parameter set is used for all evaluated scenes. The branch handling obstructed access to the target object corresponds to the shifted task structure, which is therefore explicitly encoded in the classical planner.

\paragraph{Qwen ICL baseline}
The prompt explains the planner's task, the workspace bounds, the object format, and the two skills with the 1.2~[m] proximity condition for manipulation. It lists the five admissible plan structures, including the shifted interactive-manipulation structure, followed by one demonstration per task family in Table~\ref{tab:planner_task_families} and one counter-example of an interactive-navigation scene solved incorrectly by direct navigation. The Qwen2.5-1.5B ICL baseline receives this prompt for all evaluations, including the shifted task: since it fails on the shifted task even when its structure is described and demonstrated, no reduced prompt is required. In all evaluated scenes, it outputs a single action whose skill mirrors the verb of the task instruction.

\paragraph{Frontier-model ICL baseline}

The frontier ICL baseline uses \texttt{gpt-5.5-2026-04-23} with reasoning effort \emph{low}, queried via the OpenAI API. On the ID task families, it receives the same prompt as the Qwen ICL baseline. For the evaluation on the shifted task, the shifted structure and its demonstration are removed from the prompt, leaving four plan structures and four demonstrations, so that, like our planner, the model is not informed of the shifted structure. In both cases, it generates the complete plan from the initial state in a single query, which is executed open-loop. Per-action latency is computed as the query time divided by the number of planned actions and includes network overhead; this metric was chosen due to the observed linear relationship between query latency and the number of actions generated per query. Across all held-out queries, the per-action latency is $7.0 \pm 3.4$~s (median: $5.9$~s), and a full-plan query takes $14.0 \pm 6.8$~s. A query uses on average $2{,}728$ prompt tokens and $737$ output tokens, of which $628$ are reasoning tokens; variation in reasoning length accounts for most of the latency spread.

Our planner matches or exceeds GPT-5.5 on all in-distribution task families and on the shifted task. Of its 26 interactive-navigation failures, 20 stem from geometric targets, five from selecting the wrong object, and one from insufficient clearance created by the manipulation; on long-horizon interactive navigation, eight of its 12 failures are geometric and four are skill-selection errors. On direct navigation, it tends to manipulate objects lying along the path even when sufficient clearance is available, in some cases with manipulation targets that overlap other objects. On the shifted task, it produces the correct composition in only 3/30 trajectories and completes 2/30. When the prompt additionally describes and demonstrates the shifted structure, it completes 22/30 with the correct composition in all 30, and its eight failures stem from geometric targets.

\begin{table}[t]
    \centering
    \caption{Planner evaluation task families. $\mathrm{(N)}$ and
    $\mathrm{(M)}$ denote navigation and manipulation actions, respectively.
    ID denotes task structures represented in the training distribution,
    while Shifted denotes the task composition used to evaluate
    the planner's ability to compose a novel sequence of actions.}
    \label{tab:planner_task_families}
    \begin{tabular}{lcc}
        \toprule
        \textbf{Task family} &
        \textbf{Distribution} &
        \textbf{Task composition} \\
        \midrule
        Direct navigation
            & ID
            & $\mathrm{N}$ \\

        Object manipulation
            & ID
            & $\mathrm{N}\rightarrow\mathrm{M}$ \\

        Interactive navigation
            & ID
            & $\mathrm{M}\rightarrow\mathrm{N}$ \\

        Long-horizon interactive navigation
            & ID
            & $\mathrm{N}\rightarrow\mathrm{M}\rightarrow\mathrm{N}$ \\

        Interactive manipulation
            & Shifted
            & $\mathrm{M}\rightarrow\mathrm{N}\rightarrow\mathrm{M}$ \\
        \bottomrule
    \end{tabular}
\end{table}

\subsection{Long-Horizon Interactive Manipulation}
\label{sec:LHIN}
As described in~\ref{sec:Sim_comparison_llm}, we further evaluate the planner on a longer interactive-manipulation stress test requiring the composition
$\mathrm{N}\rightarrow\mathrm{M}\rightarrow\mathrm{N}\rightarrow\mathrm{M}$ for Run 1. The purpose of this experiment is to probe how far the planner can extend the learned navigation and manipulation behaviors to a longer task structure. Across 30 trajectories, the planner completes the task only once. In six additional trajectories, it selects the correct skill sequence and satisfies the task constraints at each intermediate stage, but fails at the final manipulation where the predicted geometric target does not match the task target.

A recurrent failure pattern appears in 12 trajectories. In these cases, the planner selects the correct skill sequence,
$\mathrm{N}\rightarrow\mathrm{M}\rightarrow\mathrm{N}\rightarrow\mathrm{M}$,
and the correct object at each manipulation stage, but the geometric parameters are invalid. Specifically, the initial navigation target is unreachable under the task constraints. Rather than subsequently selecting a feasible navigation action to the target object, the planner consistently predicts a manipulation that moves the blocking object in a pulling-like behavior. This pattern is not limited to re-planning at later decision states: the same sequence is already present in the complete plan generated from the initial state. Thus, although the discrete skill and object composition is correct in these cases, the associated geometric targets make the plan infeasible.

The remaining failures consist of three trajectories in which the composed sequence is correct; however, the intermediate manipulation target does not clear a navigable path, two trajectories with invalid action format, and six trajectories in which the generated sequence instead follows the structure of the Object manipulation task family.

Based on the results of this experiment, we find that the planner is able to compose learned manipulation and navigation behaviors at the skill-level; however, to a certain extent. Failures become more frequent, and are particularly concentrated around violating action and environment constraints that are implicitly represented in the training data. Future work should investigate how to compose these learned behaviors while complying with the constraints present in the training data and across various scenes.

\subsection{Ablation and Model-Selection Analysis}
\label{sec:planner_ablation_model_selection}
\textbf{How do adaptation capacity, and action chunking affect ID and shifted-task performance?}

To that end, we examine model selection across LoRA adaptation rank and action chunking. For each run, we select the checkpoint with the lowest validation loss computed on the held-out split from which the ID test trajectories are drawn; selection is based on the loss only, not on plan success. The LoRA rank is then selected based on success on a validation set of 30 shifted-task scenes, which is disjoint from the 30 shifted-task test scenes reported in Table~\ref{tab:baselineresults_LLM}, to avoid leakage between model selection and evaluation.  %

\begin{table*}[t]

\centering
\caption{Planner ablation and model-selection results, reported as
(\# of successes)/(\# of trajectories). The shifted row reports results on the 30 validation scenes used for rank selection.}
\label{tab:planner_ablation}

\begin{tabular}{l c c c c c c c}
\toprule

\multirow{2}{*}{\textbf{Task Family}} &
\multicolumn{5}{c}{\textbf{LoRA Rank ($r$)}} &
\multicolumn{2}{c}{\textbf{No AC}} \\

\cmidrule(lr){2-6}
\cmidrule(lr){7-8}

&
\textbf{$r=32$} &
\textbf{$r=64$} &
\textbf{$r=128$} &
\textbf{$r=256$} &
\textbf{$r=512$} &
\textbf{$r=512$} &
\textbf{$r=128$} \\

\midrule

Direct navigation
    & 37/40 & 37/40 & 38/40 & 38/40 & \textbf{39/40}
    & 37/40 & \textbf{39/40} \\

Object manipulation
    & \textbf{10/10} & \textbf{10/10} & \textbf{10/10}
    & \textbf{10/10}
    & \textbf{10/10}
    & \textbf{10/10}
    & \textbf{10/10} \\

Interactive navigation
    & 98/120 & 107/120 & 109/120 & 109/120 & \textbf{112/120}
    & 107/120 & 105/120 \\

Long-horizon interactive navigation
    & 25/30 & \textbf{29/30} & \textbf{29/30} & \textbf{29/30}
    & \textbf{29/30}
    & 28/30 & 27/30 \\

\midrule

Interactive manipulation (shifted)
    & 0/30 & 0/30 & \textbf{14/30} & 6/30 & 3/30
    & 3/30 & 1/30 \\

\bottomrule
\end{tabular}
\end{table*}

First, we analyze the relationship between model success and adaptation rank across both ID and shifted trajectories. For the ID task families, performance improves up to $r=64$ and saturates thereafter. For example, the \(r=512\) model resolves three of the eleven interactive-navigation failures made by the selected \(r=128\) model, corresponding to an incorrect object selection, an incorrect geometric target, and an erroneous direct-navigation decision. In contrast, shifted-task performance does not improve monotonically with increasing rank, peaking at \(r=128\) with 14/30 successful trajectories and decreasing for both lower and higher ranks. Composition still emerges at higher ranks (11/30 and 10/30 at $r=256$ and $r=512$), but most remaining shifted-task trajectories revert to the object-manipulation sequence seen during training, whereas lower ranks additionally suffer from faulty geometric targets and erroneous output format. %

These results suggest a possible trade-off between fitting the ID task distribution and the compositional capability to the shifted task structure, although experiments across additional random seeds would be required to determine whether this behavior is systematic.  %

The model trained without action chunking achieves competitive performance on the ID trajectories; however, its performance drops sharply on the shifted interactive-manipulation task, where failures correspond to treating the scene as the familiar object-manipulation case and failing to account for the blocking object. We hypothesize that predicting the complete remaining action sequence provides a training signal that helps the planner \textit{stitch} together skills observed during training into a novel composition. The same benefit is not observed at \(r=512\), where both AC and No-AC achieve 3/30, indicating that the effect also depends on the adaptation configuration.

\subsection{Comparison with a Task-Specific Transformer}
\label{sec:planner_pretraining_ablation}

This section complements the comparison in Section~\ref{sec:Sim_comparison_llm}. Results are reported in Table~\ref{tab:baselineresults_LLM}. %

The task-specific Transformer uses a causal Transformer backbone with the same architectural components as Qwen2 (grouped-query attention, rotary position embeddings, RMSNorm, and SwiGLU feed-forward layers), with a hidden dimension of 512, 8 layers, 8 attention heads with 2 K/V heads, and a feed-forward dimension of 3072, but no pretrained weights and no text tokenizer. Instead, each semantic quantity is mapped directly into the hidden space and summed with a learned role embedding. The input prefix consists of the task (skill, object, and goal $x$, $y$, $\psi$), the robot pose, one token per scene object, and the previously executed skill. Each object token is the sum of embeddings of its identifier, type, mobility flag, size, and pose. Input geometry is continuous, with yaw encoded as $(\sin\psi, \cos\psi)$; only the output space is discretized. Each predicted action is represented by five tokens, generated in the order skill, object, $x$, $y$, $\psi$, and a separate continuation head predicts whether another action follows, which determines the plan length. Position and yaw targets are quantized at resolutions of 1~[cm] and 1~[$^\circ$], resulting in a comparable number of output classes across the geometric variables, and are trained with a cross-entropy loss. All outputs are decoded greedily.

\textit{Object pointer:} For each predicted action $j$, if $s_j = n$ (navigation), no object is selected ($i_j = -1$) and no pointer loss is applied. If $s_j = m$ (manipulation), the hidden state of the skill token, $\mathbf{h}^{\mathrm{skill}}_j$, is projected to a query, and the hidden state of each scene-object slot $o$, $\mathbf{h}^{\mathrm{obj}}_o$, to a key:
\begin{equation}
    \mathbf{q}_j = \mathbf{W}_q \mathbf{h}^{\mathrm{skill}}_j, \qquad \mathbf{k}_o = \mathbf{W}_k \mathbf{h}^{\mathrm{obj}}_o .
\end{equation}
The alignment between the query and each key is scored by a scaled dot product, normalized over the set $\mathcal{O}$ of valid object slots (padding slots are masked):
\begin{equation}
    \alpha_{jo} = \frac{\mathbf{q}_j^\top \mathbf{k}_o}{\sqrt{d}}, \qquad p(i_j = o) = \frac{\exp(\alpha_{jo})}{\sum_{o' \in \mathcal{O}} \exp(\alpha_{jo'})},
\end{equation}
where $d = 512$. The pointer is trained with a cross-entropy loss over manipulation actions,
\begin{equation}
    \mathcal{L}_{\mathrm{obj}} = - \sum_{j \,:\, s_j = m} \log p(i_j = i^\star_j),
\end{equation}
where $i^\star_j$ is the ground-truth object, and the object with the highest probability is selected at inference.

On interactive navigation, the Transformer performs closest to our planner (101/120 vs.\ 109/120), with failures mainly associated with incorrect manipulation targets. On long-horizon interactive navigation (22/30), failures arise from immediately manipulating the blocking object instead of first navigating to it, i.e., an incorrect skill composition. The largest difference is observed on object manipulation (3/10): the model predicts the expected $\mathrm{N}\rightarrow\mathrm{M}$ sequence, but the navigation target places the robot either too far from the object or close to the final task goal. One possible explanation is the comparatively limited amount of object-manipulation data in the training set. On the shifted task, the Transformer generates the intended $\mathrm{M}\rightarrow\mathrm{N}\rightarrow\mathrm{M}$ composition in all 30 test trajectories (along with the correct object to manipulate) but completes none, with the same failure mode: after the blocker is manipulated, the navigation target directs the robot toward the final task goal rather than to the target object.

The Transformer infers at $26.9 \pm 6.2$~ms per action, compared with $472.4 \pm 15.5$~ms for our planner, owing to its smaller embedding dimension and specialized tokenization.

Results of additional training runs are reported in Table~\ref{tab:seeds}.

\subsection{Qwen Architecture Trained from Scratch}
\label{sec:scratch_qwen}
To isolate the effect of pretraining from the architecture, input representation, and target discretization of the task-specific Transformer, we additionally train a model with the Qwen2.5 architecture and tokenizer from random initialization, using the same text-based inputs, outputs, and data split as our planner. Given the dataset size, we use a reduced configuration (hidden size 768, 10 layers, 12 attention heads with 2 K/V heads, feed-forward dimension 3072; approximately 201M parameters), trained for 8 epochs with a learning rate of $10^{-4}$ and otherwise the same training hyperparameters as our planner. The configuration was selected by validation loss among the tested configurations. Table~\ref{tab:scratch_qwen} compares it with our planner and the task-specific Transformer.

On the ID task families, the model trained from scratch falls well behind our planner on interactive navigation (63/120 vs.\ 109/120) and long-horizon interactive navigation (23/30 vs.\ 29/30). On the shifted task, it does not produce the correct composition in any of the 30 trajectories; instead, it manipulates the blocking object and then navigates to the manipulation goal, following the interactive-navigation structure seen during training.

This gap is expected; without pretraining, the model must learn the text-based input and output format from limited data, without prior knowledge of the vocabulary it operates on. The comparison is therefore intended to isolate the contribution of pretraining for a text-based planner rather than to serve as a competitive baseline.

\begin{table}[h]
    \centering
    \caption{Comparison of our planner with the two models trained from scratch, represented as (\# of successes)/(\# of trajectories). For the shifted task, the last row reports the number of trajectories with the correct skill composition.}
    \label{tab:scratch_qwen}
    \setlength{\tabcolsep}{2pt}
    \begin{tabular}{l c c c}
        \hline
        & \textbf{Ours} & \textbf{Qwen (scratch)} & \textbf{Transformer (scratch)} \\
        \hline
        Pretrained                  & \checkmark & --         & -- \\
        Input/output                & Text       & Text       & Structured \\
        \hline
        Direct navigation           & 38/40   & 40/40   & 35/40 \\
        Object manipulation         & 10/10   & 8/10    & 3/10 \\
        Interactive navigation      & 109/120 & 63/120  & 101/120 \\
        Long-horizon interactive nav. & 29/30 & 23/30   & 22/30 \\
        \hline
        Interactive manipulation (shifted) & 12/30 & 0/30 & 0/30 \\
        \quad correct composition   & 24/30   & 0/30    & 30/30 \\
        \hline
    \end{tabular}
\end{table}
\subsection{Additional Training Seeds}
\label{sec:seeds}
To assess sensitivity to training stochasticity, we train our planner ($r=128$) and the task-specific Transformer with two additional seeds. All runs use the same trajectory split; only weight initialization, data order, and dropout differ. Table~\ref{tab:seeds} reports the results of each run on the same evaluation trajectories as Table~\ref{tab:baselineresults_LLM}.

Success varies moderately across runs, whereas the correct skill composition is observed consistently in every run.

\begin{table}[h]
    \centering
    \caption{Planner results per training run, represented as (\# of successes)/(\# of trajectories). Runs differ only in the training seed; Run 1 corresponds to the models reported in Table~\ref{tab:baselineresults_LLM}. For the shifted task, the last row reports the number of trajectories with the correct skill composition.}
    \label{tab:seeds}
    \setlength{\tabcolsep}{2pt}
    \begin{tabular}{l c c c c c c}
        \hline
        & \multicolumn{3}{c}{\textbf{Ours}} & \multicolumn{3}{c}{\textbf{Transformer (scratch)}} \\
        \textbf{Task Family} & Run 1 & Run 2 & Run 3 & Run 1 & Run 2 & Run 3 \\
        \hline
        Direct navigation           & 38/40   & 40/40   & 40/40   & 35/40   & 34/40  & 33/40 \\
        Object manipulation         & 10/10   & 10/10   & 10/10   & 3/10    & 4/10   & 5/10 \\
        Interactive navigation      & 109/120 & 104/120 & 103/120 & 101/120 & 97/120 & 91/120 \\
        Long-horizon interactive nav. & 29/30 & 30/30   & 29/30   & 22/30   & 23/30  & 22/30 \\
        \hline
        Interactive manipulation (shifted) & 12/30 & 9/30 & 10/30 & 0/30 & 0/30 & 2/30 \\
        \quad correct composition   & 24/30   & 16/30   & 20/30   & 30/30   & 30/30  & 27/30 \\
        \hline
    \end{tabular}
\end{table}